\documentclass[lettersize,journal]{IEEEtran}
\usepackage{amsmath,amsfonts}
\usepackage{algorithmic}
\usepackage{algorithm}

\usepackage{array}
\usepackage{textcomp}
\usepackage{stfloats}
\usepackage{url}
\usepackage{verbatim}
\usepackage{graphicx}
\usepackage{cite}
\usepackage{booktabs}
\usepackage{multirow}
\usepackage{bm}
\usepackage{makecell}
\usepackage{bbding}
\usepackage{amssymb}
\usepackage{bbm}
\usepackage{mathtools} 
\usepackage{subfigure}  
\usepackage[colorlinks,
            urlcolor=black,
            linkcolor=black,      
            anchorcolor=black, 
            citecolor=black,        
            ]{hyperref}

\usepackage{verbatim}
\usepackage{etoolbox}

\begin{document}

\title{Preview-based Category Contrastive Learning for Knowledge Distillation}

\author{Muhe Ding, Jianlong Wu,~\IEEEmembership{Member,~IEEE,} Xue Dong, Xiaojie Li, Pengda Qin, Tian Gan, Liqiang Nie,~\IEEEmembership{Senior Member,~IEEE}
        % <-this % stops a space
\thanks{This work was supported by the National Natural Science Foundation of China under Grant 62376069, Grant 62236003, Grant 62176137 and Grant 62172261, in part by the Young Elite Scientists Sponsorship Program by CAST (No. 2023QNRC001).}% <-this % stops a space
    \thanks{
    Muhe Ding and Tian Gan are with the School of Computer Science and Technology, Shandong University, Qingdao 266237, China (e-mail: dmh1216380870@gmail.com, ttgantian@gmail.com).
    }
    \thanks{Jianlong Wu, Xiaojie Li, Liqiang Nie are with the School of Computer Science and Technology, Harbin Institute of Technology (Shenzhen), 518055, China (e-mail: jlwu1992@pku.edu.cn, xiaojieli0903@gmail.com, nieliqiang@gmail.com).}
    \thanks{
    Xue Dong is with the School of Software, Shandong University, Jinan 250101, China (e-mail: dongxue.sdu@gmail.com).
    }
    \thanks{Pengda Qin is with Alibaba Group, Beijing 100102, China (e-mail: pengda.qpd@alibaba-inc.com).}
}

\markboth{IEEE TRANSACTIONS ON CIRCUITS AND SYSTEMS FOR VIDEO TECHNOLOGY}
{Ding \MakeLowercase{\textit{et al.}}: Preview-based Category Contrastive Learning for Knowledge Distillation}
%\markboth{IEEE Transactions on Image Processing}%


\maketitle

\begin{abstract}
Knowledge distillation is a mainstream algorithm in model compression by transferring knowledge from the larger model~(teacher) to the smaller model~(student) to improve the performance of student. Despite many efforts, existing methods mainly investigate the consistency between instance-level feature representation or prediction, which neglects the category-level information and the difficulty of each sample, leading to undesirable performance. To address these issues, we propose a novel preview-based category contrastive learning method for knowledge distillation~(PCKD). It first distills the structural knowledge of both instance-level feature correspondence and the relation between instance features and category centers in a contrastive learning fashion, which can explicitly optimize the category representation and explore the distinct correlation between representations of instances and categories, contributing to discriminative category centers and better classification results. Besides, we introduce a novel preview strategy to dynamically determine how much the student should learn from each sample according to their difficulty. Different from existing methods that treat all samples equally and curriculum learning that simply filters out hard samples, our method assigns a small weight for hard instances as a preview to better guide the student training. Extensive experiments on several challenging datasets, including CIFAR-100 and ImageNet, demonstrate the superiority over state-of-the-art methods.
\end{abstract}


\section{Introduction} \label{introduction}
\IEEEPARstart{C}{onvolutional} neural networks~(CNNs) show superiority in many computer vision tasks, such as image classification~\cite{krizhevsky2012imagenet,he2016deep,DBLP:journals/tcsv/MaPCZDT23,DBLP:journals/tcsv/YangJLLYLCLH23}, object detection~\cite{girshick2014rich,redmon2016you,chen2023sd,DBLP:journals/tcsv/XieHGWZ24}, and semantic segmentation~\cite{long2015fully,shu2021channel,DBLP:journals/tcsv/WuFLLL24,lu2024self}. 
	A number of attempts have been made to devise a more complex architecture for CNNs in a bid to achieve better accuracy~\cite{simonyan2014very,he2016deep}. However, such complex networks inevitably cost more parameters and computations, making them difficult to deploy in real applications. Thus, recent years have witnessed a growing trend of model compression~\cite{bucilua2006model,denton2014exploiting,han2015learning,wu2016quantized,cheng2018model}, which aims to simplify the network architecture and decrease the network parameters. Knowledge Distillation~(KD)~\cite{hinton2015distilling} is the favored method of model compression. The crucial idea of KD is to distill the knowledge of the complex networks~(i.e., the teacher)~into a more lightweight network~(i.e., the student). With the guidance of the teacher, the student is able to achieve similar performance as the teacher but with easier architecture and fewer parameters.

\begin{figure}[!t]
\centering
\subfigbottomskip=2pt 
	\subfigcapskip=-5pt 
	\subfigure[Illustration of existing knowledge distillation methods.]{
        \includegraphics[width=3.5in]{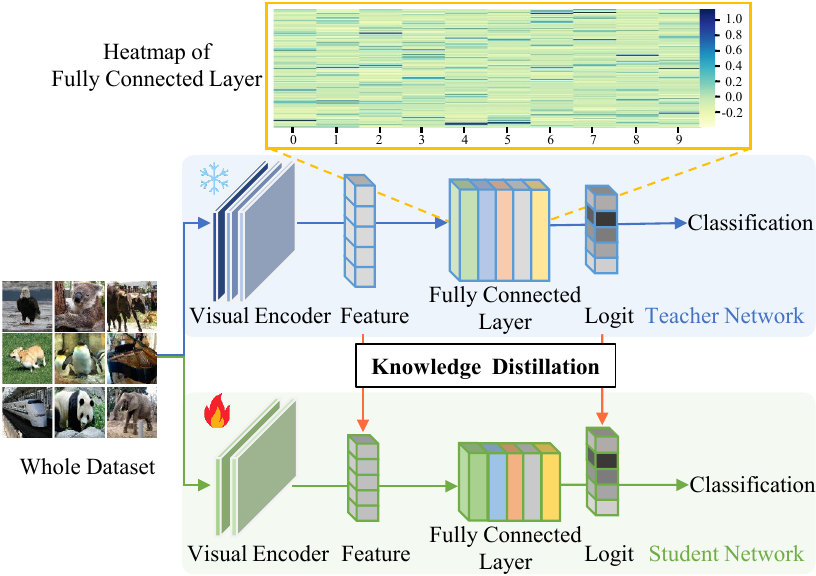}}
    \subfigure[Illustration of preview-based learning strategy.]{
        \includegraphics[width=3.5in]{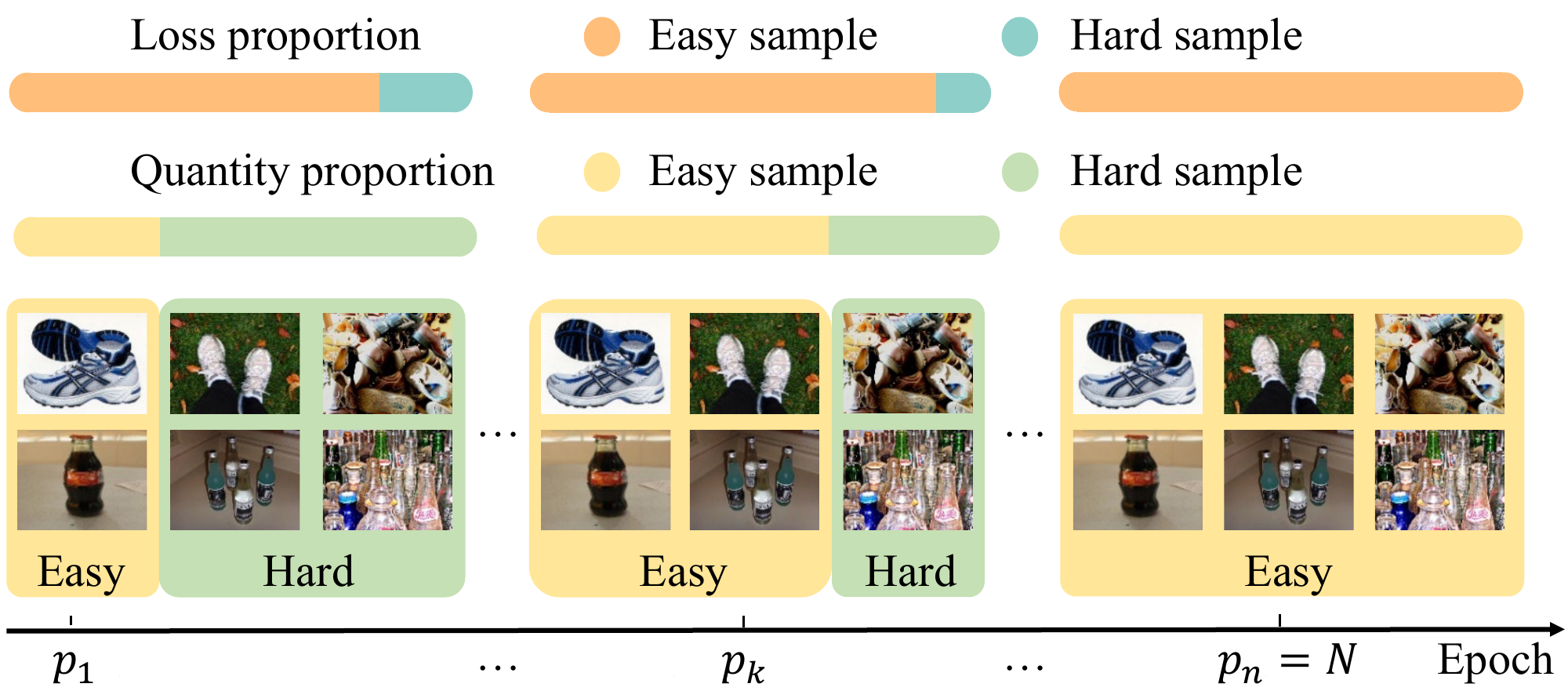}}\caption{The motivation of our proposed method. (a) Existing knowledge distillation methods mainly transfer knowledge of features and logits, ignoring the category-level information in the parameters of the fully connected layer.
			(b) Illustration of our proposed preview-based learning strategy. It dynamically adjusts the difficulties of input instances and gradually increases their learning weights during the training.}
        \vspace{-4mm}
\label{motivation}
\end{figure}

    Generally speaking, the key of KD approaches is to extract sufficient knowledge from a teacher network to guide a student network.  
	For example, Hinton~et~al.~\cite{hinton2015distilling} first propose to distill the knowledge from teachers' output logits to student through the Kullback-Leibler divergence. Several approaches~\cite{romero2014fitnets,zagoruyko2016paying} further introduce the distillation in the feature maps at the intermediate layer between the teacher and student with the Euclidean distance. However, these methods only distill instance-level consistent information. Considering the structural information in the feature space of the teacher, some researches~\cite{tian2019contrastive, zhu2021complementary} introduce contrastive learning into the knowledge distillation to learn discriminative feature representations.

	Despite their effectiveness, current approaches suffer from the following issues. 
	1) Existing methods mainly make the student learn the results of the teacher while ignoring teaching the student how the teacher operates to derive the results. In other words, they mainly utilize the teachers' results~(e.g., the learn features or logits) to guide the student while ignoring the inherent knowledge~(e.g., the architecture and parameters) of the teacher. In fact, as can be seen from Fig.~\ref{motivation}~(a), the output logit of an instance is derived from the feature of the last convolutional layer after a fully connected layer. To be more specific, the weight matrix of this fully connected layer operates on perceiving the similarity between the instance and each category, and outputting the probabilities of the instance belonging to all categories. Therefore, we consider the weight matrix as the category center to classify the instance. To better understand the category centers, we train a state-of-the-art teacher network WRN-40-2~\cite{zagoruyko2016wide} on CIFAR-100 dataset~\cite{krizhevsky2009learning} and visualize 10 category centers in the heatmap of Fig.~\ref{motivation}~(a), where each column indicates a certain category center and the darker color indicates the bigger value. As can be seen, each category center is different and representative. Inspired by this, we argue that the category centers show how the teacher derives the results and also need to be distilled to the student. 2) Existing approaches make the student learn all the knowledge of the teacher without distinguishing the difficulty of the knowledge. In fact, from Fig.~\ref{motivation}~(a), we can see that some input images are with a single object and clear background, while others are with multiple objects and complex backgrounds. The student network has simple architecture and fewer parameters, like a pupil that cannot accept all knowledge from the teachers at the beginning, especially the hard knowledge. Thus, it is necessary to design a new learning strategy that enables the student to learn from easy knowledge and acquire hard knowledge gradually. 

    To solve the aforementioned problems, we propose a novel \textbf{P}review-based \textbf{C}ategory contrastive learning for \textbf{K}nowledge \textbf{D}istillation, named PCKD.
	It consists of two key components: category contrastive learning for knowledge distillation~(CKD) and preview-based learning strategy, which are able to teach the student how to operate and help the student learn better, respectively. As for category contrastive learning for knowledge distillation, we make the student learn the results~(i.e., feature and logits) of the teacher, the operation~(i.e., category centers) of the teacher, and their correlations. Besides, to make the category center more representative and discriminative for the corresponding category, we introduce a contrastive loss to make the feature closed to its corresponding category center while far from other class centers. As for preview-based learning strategy, shown in Fig.~\ref{motivation}~(b), we propose it to help students better learn knowledge taught by the teacher, whose philosophy is adding a smaller weight to those hard instances as a preview.
    To be more specific, we determine the difficulty of an instance based on its classification loss of the student network. Then the weights can be dynamically adjusted according to their difficulty during training, which helps the student network learn progressively.
	Experimental results on commonly-used challenging datasets show the effectiveness of the proposed method. We also perform extensive ablation analysis to demonstrate the superiority of PCKD.
	
	Our contributions can be summarized as three points:
	\begin{itemize}
		\item  We propose a novel Preview-based Category contrastive learning for Knowledge Distillation~(PCKD), which transfers the knowledge in all the results, operations of the teacher, and their correlations to the student in a contrastive learning manner. In this way, the student can learn how the teacher makes predictions, and get more discriminative category centers as well as achieve better performance.
		\item We introduce a novel preview-based learning strategy for the student, which adds adjustable weights to different instances based on their difficulty score during the training. This strategy enables the student to learn from the teacher more progressively and effectively.
		\item We conduct extensive experiments and ablation studies on different datasets~(i.e., CIFAR-100~\cite{krizhevsky2009learning}, ImageNet~\cite{deng2009imagenet}, STL-10~\cite{coates2011analysis}, and TinyImageNet~\cite{deng2009imagenet}) with different network architectures. Experimental results demonstrate our superiority over existing state-of-the-art methods.
	\end{itemize}

	The rest of this paper is organized as follows. Section~\ref{a} briefly reviews the related work about knowledge distillation, contrastive learning, and curriculum learning. In~Section~\ref{b}, we elaborately introduce the proposed preview-based category contrastive learning for knowledge distillation. Section~\ref{c} presents experimental results and analyses, followed by the conclusion in Section~\ref{d}.

\section{Related Work}\label{a}
	\subsection{Knowledge Distillation}
	The foundational concept of knowledge distillation was introduced by KD~\cite{hinton2015distilling} where it fits the logits of the teacher and student model by reducing the KL divergence to transfer \textit{dark knowledge}, which can significantly improve the performance of the student network without introducing extra parameters.
	Since then, numerous approaches have been proposed to explore various kinds of knowledge and narrow the gap between the teacher and the student. 
    According to the type of knowledge used for distillation, existing methods can be mainly classified into three categories, i.e., response-based KD, feature-based KD, and relation-based KD~\cite{gou2021knowledge}.
	
	Response-based KD directly transfers the prediction of the penultimate output layer~(i.e., logits) as knowledge and wants the student model to imitate the output of the teacher model directly. The most typical is the classic KD method~\cite{hinton2015distilling}.
    DKD~\cite{DBLP:conf/cvpr/ZhaoCSQL22} revealed the classic KD method is a coupling formula and proposed decoupled knowledge distillation.
    LSKD~\cite{li2023boosting} investigated the reason rendering knowledge distillation and label smoothing to exert distinct effects on model’s potential ability in sequential knowledge transferring.
    MLD~\cite{DBLP:conf/cvpr/JinWL23} explored stronger logits distillation through multi-level prediction alignment.
	Feature-based KD aims to transfer the feature of the intermediate layer as knowledge. FitNets~\cite{romero2014fitnets} proposed to learn the middle layer of the network as knowledge. 
    AT~\cite{zagoruyko2016paying} introduced the attention mechanism and forced the student network to mimic the spatial maps of the teacher model. SemCKD~\cite{wang2022semckd} exploited intermediate knowledge by semantic calibration and feature-map transfer across multilayers.
	Different from the single instance-based knowledge, relation-based knowledge regards the relations and structures among various data samples or layers as knowledge. RKD~\cite{park2019relational} and CC~\cite{peng2019correlation} transferred the structured feature correlation among samples as knowledge. 
	FSP~\cite{yim2017gift} established the Gram matrix for different feature maps to transfer knowledge.  
	LKD~\cite{li2020local} distilled local correlation consistency. ICKD~\cite{liu2021exploring} mined structural information of inter-channel correlation. ReviewKD~\cite{DBLP:conf/cvpr/Chen0ZJ21} studied the factor of connection path cross levels between teacher and student networks.
    EKD~\cite{DBLP:journals/tcsv/ZhangZLZG22} improved the transfer effectiveness of teacher knowledge by evolutionary KD.
    DCGD~\cite{DBLP:journals/tcsv/XuWXLY24} employed divide and conquer group distillation to transfer teachers' knowledge by grouping target tasks into small-scale subtasks and designing multi-branch networks.
    Recently, online knowledge distillation has become popular. DCCL~\cite{DBLP:journals/tcsv/SuLZYXWL23} proposed deep cross-layer collaborative learning network for online knowledge distillation, and FFSD~\cite{DBLP:journals/tnn/LiLWWTSJ23} presented feature fusion and self-distillation.
	Compared with the original student network and vanilla KD, the above methods achieve good improvement, but the instance-level feature correspondence and relation-based knowledge are not sufficient enough, and our PCKD can distill both results and operations of the teacher in a contrastive way to get discriminative category representation and relation.
	
	\subsection{Contrastive Learning}
	Contrastive learning~\cite{chen2020simple,he2020momentum,wu2018unsupervised,grill2020bootstrap} has become a mainstream method in self-supervised learning, especially in the fields of computer vision and machine learning~\cite{jaiswal2020survey}.   
	By constructing positive and negative sample pairs first, contrastive learning minimizes the distance of positive pairs to learn consistent features and enlarges the distance of negative pairs to realize the distinction between different categories.
	SimCLR~\cite{chen2020simple} constructed positive samples by data augmentation and simply regarded other samples within the same mini-batch as negative samples, which generally requires a large batch size to maintain the proportion of negative samples.
	MoCo~\cite{he2020momentum} proposed the momentum encoder to construct negative samples and achieved good results while maintaining a small batch size through the momentum update method. BYOL~\cite{grill2020bootstrap} used two networks, the online network and the target network, to learn and influence each other.
	Besides the above unsupervised setting,
	SupCon~\cite{khosla2020supervised} further utilized label information and extended the self-supervised contrastive method to the fully-supervised situation on the basis of SimCLR.  
	Recently, several methods incorporated contrastive learning into knowledge distillation to learn better structural knowledge.
	For example, CRD~\cite{tian2019contrastive} maximized a lower-bound to the mutual information between the teacher and student representation based on the contrastive setting. 
	CRCD~\cite{zhu2021complementary} further transferred the sample representation and inter-sample relations as structured knowledge. 
	SSKD~\cite{xu2020knowledge} presented to use contrastive learning on transformed images as the self-supervision pretext tasks, which can exploit the self-supervised contrastive relationships.
	HSAKD~\cite{DBLP:conf/ijcai/YangACX21} added several auxiliary classifiers to the intermediate feature maps to generate self-supervised knowledge, but the network of this method becomes larger and violates the purpose of model compression.
	
	\subsection{Curriculum Learning}
	The idea of curriculum learning is inspired by the human learning process. Human usually learns easy knowledge first and then gradually understands more complicated concepts. The concept of curriculum learning is first proposed by~\cite{bengio2009curriculum} and it has been widely adopted for image classification~\cite{tang2012self,gong2016multi} and object detection~\cite{zhang2019leveraging}. Since the curriculum is predetermined by prior knowledge, Kumar~et~al.~\cite{kumar2010self} presented the self-paced learning algorithm to generate the curriculum by the learner itself. Tang~et~al.~\cite{tang2012self} proposed to adaptively choose easy samples in each iteration for dictionary learning. Combining the merits of prior knowledge and self-paced learning, Jiang~et~al.~\cite{jiang2015self} came up with a learning paradigm called SPCL. 
    More recently, Graves~et~al.~\cite{graves2017automated} proposed an algorithm for automatically selecting syllabus to improve the efficiency of curriculum selection. Fan~et~al.~\cite{fan2018learning} incorporated teacher guidance into curriculum learning in which two intelligent agents~(called student model and teacher model) interact with each other. At present, curriculum learning is rarely used in knowledge distillation. Zhu~et~al.~\cite{zhu2021combining} combined the knowledge distillation and the curriculum learning to achieve good results in dialogue generation. Tudor~et~al.~\cite{tudor2016hard} presented the human response time for estimating image difficulty. Xiang~et~al.~\cite{xiang2020learning} presented the LFME framework, including self-paced knowledge distillation and curriculum instance selection, which is specially designed for long-tailed classification and unsuitable for general knowledge distillation. Instead of the classic curriculum learning, we further present a novel preview strategy to assist the student model to evaluate the difficulty level of samples, so that it can assign a small weight for these hard examples at the beginning and accept the guidance of the teacher model more easily, which is also consistent with the human learning process.
	
	\begin{figure*}[!t]
		\centering
		\includegraphics[width=1.02\textwidth]{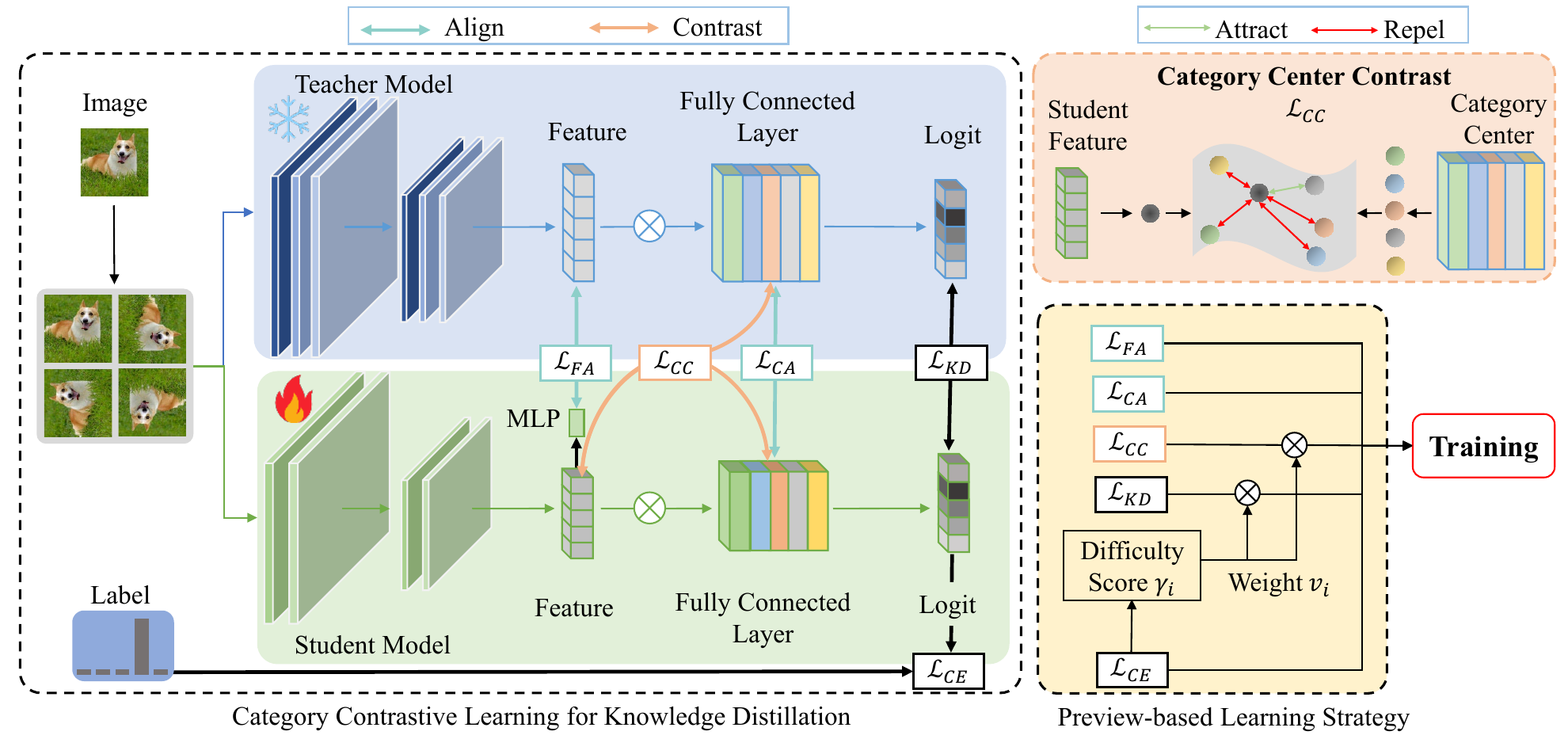}
		\caption{The overall framework of our proposed PCKD. We first augment samples, extract features and perform feature alignment~($\mathcal{L}_{FA}$), category center alignment~($\mathcal{L}_{CA}$), and category center contrast~($\mathcal{L}_{CC}$). 
		Then our preview strategy can assign dynamic weights to each sample based on its difficulty score.}
		\label{Figure_2}
	\end{figure*}
	
\section{Methodology}\label{b}
	
	In this section, we first briefly introduce the basic framework of knowledge distillation. Then we present the details of our proposed category contrastive learning strategy and preview-based learning strategy. Finally, we conclude the overall framework of our approach~(PCKD). 
	
	\subsection{Preliminary: Vanilla Knowledge Distillation}
	
	Let $\mathcal{X}=\{x_i\}^N_{i=1}$ denote a set of training examples, whose corresponding label set is $\mathcal{Y}=\{y_i\}^N_{i=1}$, and $N$ is the number of samples. 
	For each sample $x_i$, we denote logits of teacher and student models as
	$z^T(x_{i})$ and $z^S(x_{i})$, respectively. Given a pre-trained teacher model $T$, KD aims to learn a better student model $S$ by transferring the knowledge (e.g., features or logits) from teacher model. 
	For example, Vanilla KD~\cite{hinton2015distilling} minimizes the Kullback-Leibler divergence between the logits output to mimic the output of the teacher network. 
	The basic loss function is defined as:
	\begin{equation}
		\label{e1}
		\mathcal{L}_{KD}=\frac{1}{N} \sum_{x_i \in \mathcal{X}} \text{KL}(\rho(\frac{z^T(x_{i})}{\tau}), \rho(\frac{z^S(x_{i})}{\tau})),
	\end{equation}
	where $\tau$ is the weight factor as temperature and $\rho(\cdot)$ is the softmax function.
	
	\subsection{Category Contrastive Learning}
	\label{section_category_contrastive_learning}
	This subsection details the Category contrastive learning for Knowledge Distillation~(CKD). Unlike existing methods, CKD makes the student learn from three aspects, including the results (i.e., feature and logits) and operation (i.e., category center) of the teacher, as well as their correlation. Following the settings of self-supervised learning for knowledge distillation~\cite{xu2020knowledge}, we copy the image of the training sample $x_i$ three times to new samples $x_{i1}, x_{i2}$, $x_{i3}$ and rotate them by $90^{\circ}$, $180^{\circ}$, and $270^{\circ}$, respectively. This process triples the training set, thereby introducing additional positive and negative samples for contrastive learning.

	\subsubsection{Feature Alignment}
	\label{section_fa}
	Denote $f^T(x_{ij})$ and $f^S(x_{ij})$ as the features of the $j$-th copy~($j=0,1,2,3$) of the $i$-th sample in the teacher and student networks, respectively.
	The feature alignment aims to make the feature in the student~$f^S(x_{ij})$ mimic the feature of the teacher~$f^T(x_{ij})$.
	Since there is a wide gap between these two features, referring to~\cite{grill2020bootstrap} and~\cite{ding2020multi}, we utilize a multilayer perceptron~$g$ with one hidden layer over the student feature~$f^S(x_{ij})$ to encode the higher-order dependency of the teacher network. We make the encoded student feature~$g\left(f^S(x_{ij})\right)$ similar to the teacher feature~$f^T(x_{ij})$ through the following loss function:
	\begin{equation}
		\label{e2}
		\mathcal{L}_{FA} =\sum_{i,j}\left\|\overline{g}\left(f^{S}(x_{ij})\right)-\overline{f^{T}}(x_{ij})\right\|_{2}^{2},
	\end{equation}
	where $\overline{g}\left(f^{S}(x_{ij})\right)$ and $\overline{f^{T}}(x_{ij})$ are the normalized encoded student feature and the teacher feature, respectively. To be more specific, $\overline{g}\left(f^{S}(x_{ij})\right)= \frac{g\left(f^{S}(x_{ij})\right)}{\left\|g\left(f^{S}(x_{ij})\right)\right\|_{2}}$, and $\overline{f^{T}}(x_{ij})=\frac{f^{T}(x_{ij})}{\left\|f^{T}(x_{ij})\right\|_{2}}$, where $\left\|\cdot\right\|_2$ refers to the $\ell_2$-norm of the vector.

	\begin{figure}[!t]
		\centering
		\includegraphics[width=3.5in]{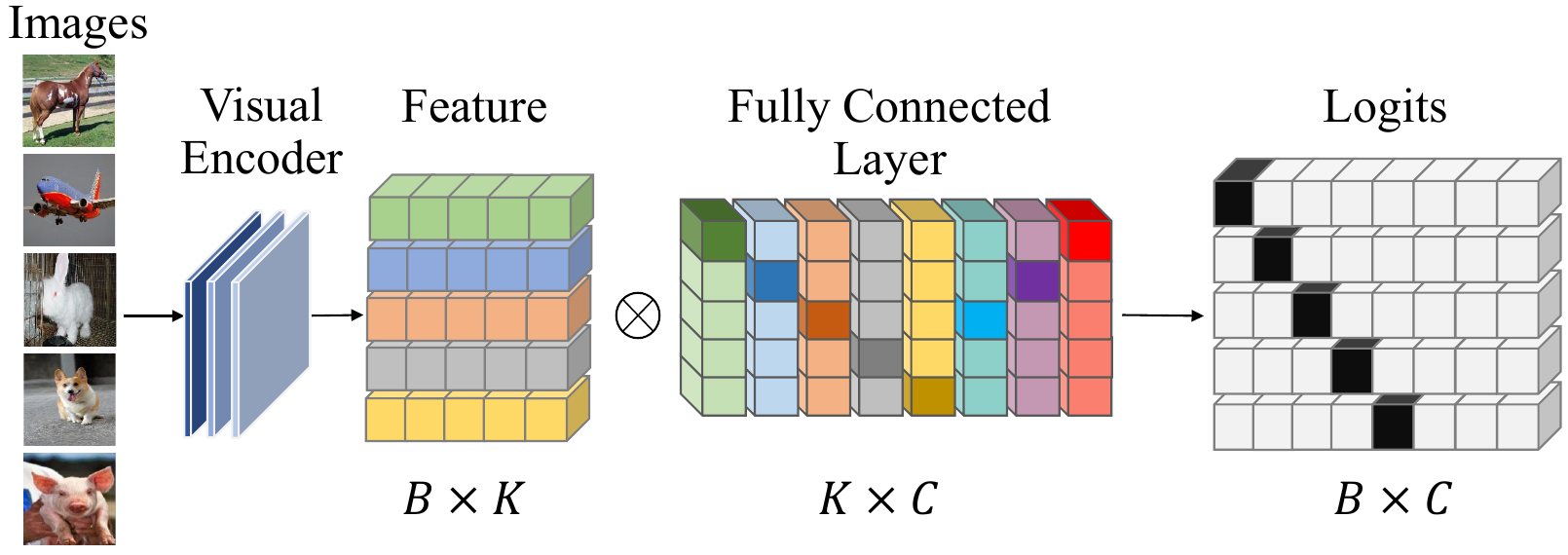}
		\caption{Illustration of multiplication between the feature vectors and weight matrix in the fully connected layer (best viewed in color). Each column vector in the weight matrix is regarded as a category center, representing a specific category.
			$B$ is the batch size, $K$ is the feature size, and $C$ denotes the total number of categories. }
		\label{center}
	\end{figure}

	\subsubsection{Category Center Alignment} 
	\label{section_ca}
	In CNNs, the teacher derives the feature of an input instance through multiple convolutional layers and pooling layers, and then adopts a fully connected layer to map the feature to its logit for the classification. 
	Specifically, as shown in Fig.~\ref{center}, the weight matrix of the fully connected layer operates on perceiving the similarity between the instance feature and each category, whereby outputs the probabilities of the instance belonging to all categories. We term each column of the weight matrix as a category center that represents the properties of a specific category. Accordingly, we distill the knowledge in the category centers to the student, enabling the student to learn how the teacher operates to classify the instance.
	Technically, we make the student learn the category centers of the teacher through the following loss function:
	\begin{equation}
		\label{e3}
		\mathcal{L}_{CA}=\left\|W^{T}-W^{S}\right\|^{2}_2,
	\end{equation}
	where $W^T$ and $W^S$ are category centers (i.e., the weight of the fully connected layer) of the teacher and student networks, respectively. 
	
	\subsubsection{Category Center Contrast}
	\label{section_cc}
	As aforementioned, the category center represents the properties of a certain category, so it is expected to be representative and discriminative for the corresponding category. Therefore, inspired by the contrastive learning~\cite{khosla2020supervised, xu2020knowledge}, 
	we enforce the feature of the student network~$f^S(x_{ij})$ to be as similar as the corresponding category centers of both the teacher network~$W^T_{y_i}$ and student network~$W^S_{y_i}$ while far from other category centers, where $y_i$ is the ground-truth category of the $i$-th sample. Technically, we estimate the similarity between the sample feature and category center with their cosine distance. Then we define the following contrastive loss:
	\begin{equation}
		\label{e4}
		\begin{aligned}
			&\quad \mathcal{L}_{CC}= \mathcal{L}_{CC}^{T}+\mathcal{L}_{CC}^{S}, \\
			&\left\{
			\begin{aligned}
				\mathcal{L}_{CC}^{T}&=-\sum_{i,j}\log \frac{\exp \left(f^{S}\left(x_{ij}\right) \cdot W_{y_i}^{T} / \tau\right)}{\sum_{c=1, c\neq y_{i}}^{C} \exp \left(f^{S}\left(x_{ij}\right)  \cdot W_c^{T} / \tau\right)}, \\
				\mathcal{L}_{CC}^{S}&=-\sum_{i,j}\log \frac{\exp \left(f^{S}\left(x_{ij}\right) \cdot W_{y_i}^{S} / \tau\right)}{\sum_{c=1, c\neq y_{i}}^{C} \exp \left(f^{S}\left(x_{ij}\right) \cdot W_{c}^{S} / \tau\right)}, \\
			\end{aligned}
			\right.
		\end{aligned}
	\end{equation}
	where $f^S(x_{ij})$ is the student feature for the $j$-th copy of the $i$-th sample, and $x_{i0}=x_i$. $C$ is the number of all categories. $\tau$ is the temperature parameter. 
	
	By minimizing the above loss function, the distance between the student network feature $f^S(x_{ij})$ and its corresponding category centers~($W_{y_i}^S$ and $W_{y_i}^T$) is forced to be smaller than the distance from other unmatched category centers~($W_{c}^{S}$ and $W_{c}^{T}$), which can also explicitly guide the discriminative category centers learning. Accordingly, the category centers as well as the learned student network features can be representative and discriminative.

	\begin{figure}[!t]
		\centering
		\includegraphics[width=3.5in]{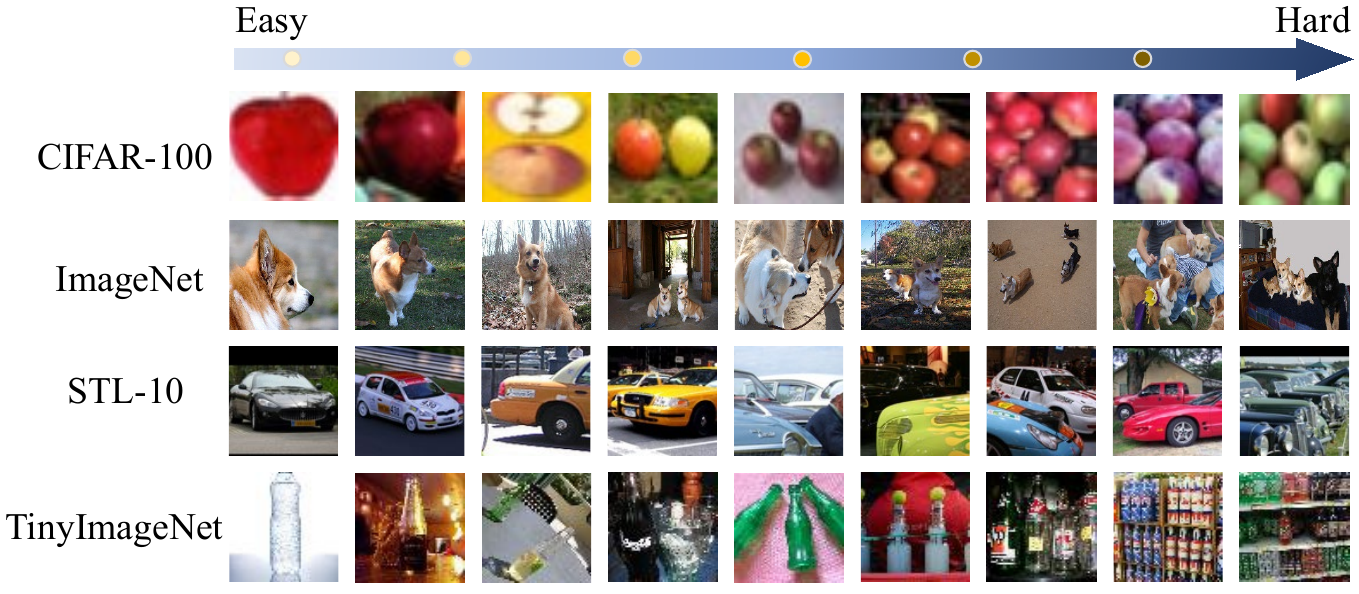}
		\caption{Sample images of different difficulties on various datasets. The difficulty score of these images increases from left to right.}
		\label{difficulty}
	\end{figure}
	
	\subsection{Preview-based Learning Strategy}
	\label{section_preview_learning_strategy}
	In the practical computer vision tasks, there exists a large variance among different images~\cite{soviany2020image}. As shown in Fig.~\ref{difficulty}, the images on the left are with a single object and clear background, which are easy to recognize. In contrast, images on the right with multiple objects and cluttered backgrounds are harder to recognize.
	In this case, it is difficult for the student network with simple architecture and few parameters to accept all knowledge from the teacher network directly. 
	Towards this issue, the curriculum learning methods~\cite{bengio2009curriculum,wang2021survey} propose to pre-calculate the sample difficulty and then make the student learn from the easy samples as well as filter out these hard samples.
	Nevertheless, in real-world scenarios, teachers usually teach students knowledge of this lesson and encourage students to preview the difficult knowledge of the next lesson after class. This practice aids students in gaining a deeper understanding of intricate lesson knowledge.
	Inspired by this, we propose a new preview-based learning strategy to make the student learn the simple knowledge as well as dynamically preview the difficult knowledge during the training. 
    Preview-based learning strategy has two crucial problems: (1) how to calculate the difficulty of each sample; (2) how to adaptively preview the hard knowledge within the training process.
	
	\textbf{Difficulty Score Calculation. }
	Referring to~\cite{kumar2010self}, if the student network can correctly classify a sample, we infer it is an easy sample, and vice versa for a hard sample. Technically, the cross-entropy loss of a sample indicates how much its prediction is similar to its label, and it can be regarded as the sample difficulty. Accordingly, we define the difficulty score $\gamma_i$ of sample~$x_i$ as follows:
	\begin{equation}
		\label{e5}
        \gamma_{i}=\frac{\mathcal{L}_{CE}\left(x_{i}\right)}{\frac{1}{\vert \mathcal{X}_b \vert }\sum_{x_j \in \mathcal{X}_b} \mathcal{L}_{CE}(x_j)},
	\end{equation}
	where $\mathcal{L}_{CE} (x_i)$ is the cross-entropy loss of sample $x_i$, $\mathcal{X}_b$ contains all samples in the current batch, and $\vert\mathcal{X}_b\vert$ denotes the number of samples in one mini-batch. 
	We utilize the mean difficulty of all samples in the current batch to normalize the sample difficulty. 
	Note that one can choose $\mathcal{X}_b$ to be the whole training samples, but this is not practical during training due to a large number of images and high computations of calculating the cross-entropy losses.
	For this score, the higher value indicates that this sample is more difficult to learn.

	\textbf{Learning Weight of the Sample. }
	In the context of our preview-based learning strategy, we make the student not only concentrate on the easy samples but also preview the hard samples. In other words, in one batch of the training, preview-based learning strategy pays more attention to making the student correctly classify the easy samples while less attention to the hard samples. To fulfill this, we introduce an auxiliary weight $v_i$ to each sample as its attention during the training as follows:
	\begin{equation}
		\label{e6}
		v_{i}=\left\{\begin{array}{cl}
			1 & \text { if } \gamma_i \leq \lambda, \\
			\exp \left(-\gamma_{i}^{2}\right) & \text { if } \gamma_i>\lambda,
		\end{array}\right.
	\end{equation}
	where $\gamma_{i}$ is the difficulty score defined in Eq.~(\ref{e5}). $\lambda\geq 1$ is the threshold to determine whether the sample is easy or hard, i.e., the sample~$x_i$ is easy if $\gamma_{i} \leq \lambda$, and vice versa. According to Eq.~(\ref{e6}), in each training batch, the learning weights of easy samples are assigned to~$1$, while those of hard samples are smaller than~$e^{-1}=0.367$.
	
	Besides, along with the training, the student network accumulates more knowledge from the teacher, and it is able to learn more hard knowledge. We thus dynamically raise threshold~$\lambda$ at each training epoch and make the student learn more hard knowledge. Technically, following the study~\cite{kumar2010self}, we define the threshold $\lambda$ controlled by the training epoch~$t$ as follows:
	\begin{equation}
		\label{e7}
		\lambda=(1+\varepsilon)^{t},
	\end{equation}
	where $\varepsilon$ is a hyperparameter to control the growth rate. As shown in Fig.~\ref{focal_figure}, along with the increase of $t$, $\lambda$ becomes larger, and almost all hard samples will become easy samples at the end.
	
	\subsection{Preview-based Category Contrastive Learning for Knowledge Distillation}
	Ultimately, based on the category contrastive learning and preview-based learning strategy in subsections~\ref{section_category_contrastive_learning} and~\ref{section_preview_learning_strategy}, respectively, the final loss function of the proposed preview-based category contrastive learning for knowledge distillation~(PCKD) is defined as follows:
	% \begin{equation}
	% 	\label{e8}
	% 	\begin{aligned}
	% 		\mathcal{L}_{PCKD} =  & \sum_{i=1}^{N}\big(\mathcal{L}_{CE}(x_i)+\alpha  v_{i} \mathcal{L}_{KD}(x_i) +\beta_{cc} v_{i}  \sum_{j=0}^3\mathcal{L}_{CC}\left(x_{ij}\right)
	% 		  \\&+\beta_{fa} \sum_{j=0}^3\mathcal{L}_{FA}(x_{ij})  \big) + \beta_{ca} \mathcal{L}_{CA},
	% 	\end{aligned}
	% \end{equation}
    \begin{equation}
		\label{e8}
		\begin{aligned}
			\mathcal{L}_{PCKD} =  & \sum_{i=1}^{N}\Big( \alpha v_{i}\mathcal{L}_{KD}(x_i) +\beta_{cc} v_{i}\sum_{j=0}^3\mathcal{L}_{CC}\left(x_{ij}\right)+
			  \\& \mathcal{L}_{CE}(x_i)+\beta_{fa} \sum_{j=0}^3\mathcal{L}_{FA}(x_{ij})  \Big) + \beta_{ca} \mathcal{L}_{CA},
		\end{aligned}
	\end{equation}
	where $\mathcal{L}_{CE}$ is the cross-entropy loss in the student network, which enforces the student to make the correct classification. $\mathcal{L}_{KD}$ is the basic loss of the knowledge distillation defined in Eq.~(\ref{e1}). $\alpha$ is the trade-off parameter and we set it to $1$ in experiments. $\mathcal{L}_{CC}$ is the contrastive loss defined in Eq.~(\ref{e4}). $\mathcal{L}_{FA}$ and $\mathcal{L}_{CA}$ are the losses of the feature and category center alignments defined in Eqs.~(\ref{e2}) and (\ref{e3}), respectively. $\beta_{cc}$, $\beta_{fa}$, and $\beta_{ca}$ are the trade-off parameters to balance the corresponding terms~(we will discuss the effects of these parameters in subsection~\ref{section_parameter_analysis}). Since the learning weight $v_i$ is designed to control the attention of the student to classify samples~$x_i$ correctly, we only add it into the losses that are utilized for the classification, i.e., $\mathcal{L}_{KD}$ and $\mathcal{L}_{CC}$. More discussions about which losses should be combined with learning weight $v_i$ from preview-based learning strategy are in subsection~\ref{ Preview-based}.
    
	To give a clear illustration of our proposed PCKD, we summarize the training process in Algorithm~\ref{alg1}. The overall framework can also be found in Fig.~\ref{Figure_2}.
	
	\begin{algorithm}[t]
		\caption{The Algorithm for PCKD.}
		\begin{algorithmic}
			\STATE \textbf{Input:} Pre-trained teacher model $T(\theta_t)$ and dataset $\{x_i\}_{i=1}^{N}$.
			\STATE \textbf{Output:} Student model $S(\theta_s)$.
			\STATE Initialize the student model with $\theta_s$ and freeze the teacher model with fixed parameters $\theta_t$.
			\STATE $\widetilde{x}$ = augmentation$(x)$.
			\STATE \textbf{for} sampled mini-batch $\{\widetilde{x}_i\}_{i=1}^{N}$
			\textbf{do}  
			\STATE \hspace{0.5cm} \textbf{Step 1:} Calculate feature alignment loss $\mathcal{L}_{FA}$
			\STATE \hspace{0.5cm} by Eq.~(\ref{e2});
			\STATE \hspace{0.5cm} \textbf{Step 2:} Compute category center alignment loss $\mathcal{L}_{CA}$ 
			\STATE \hspace{0.5cm} by Eq.~(\ref{e3});
			\STATE \hspace{0.5cm} \textbf{Step 3:} Calculate category center contrast loss $\mathcal{L}_{CC}$ 
			\STATE \hspace{0.5cm} by Eq.~(\ref{e4});
			\STATE \hspace{0.5cm} \textbf{Step 4:} Calculate cross-entropy loss $\mathcal{L}_{CE}$ and KD 
			\STATE \hspace{0.5cm} loss $\mathcal{L}_{KD}$ by Eq.~(\ref{e1});
			\STATE \hspace{0.5cm} \textbf{Step 5:} Compute the difficulty score of each sample  
			\STATE \hspace{0.5cm} $\gamma_i$ by Eq.~(\ref{e5}) and assign weight $v_i$ by Eq.~(\ref{e6});
			\STATE \hspace{0.5cm} \textbf{Step 6:} Calculate the total loss $\mathcal{L}_{PCKD}$ by Eq.~(\ref{e8});
			\STATE \hspace{0.5cm} \textbf{Step 7:} Update the parameters of student model $\theta_s$:
			\STATE \hspace{1.7cm} $\theta_s=\arg\min_{\theta_s}\mathcal{L}_{PCKD}$.
			\STATE \textbf{end for}
		\end{algorithmic}
		\label{alg1}
	\end{algorithm}

\section{Experiments}\label{c}
	In this section, we first validated the performance of our proposed method on four challenging image classification datasets under various combination of backbones.
	Then we conducted experiments to evaluate the transferability of learned feature representations. 
	The ablation studies are also conducted to verify the effectiveness of each module in our method. 
	Besides, we performed a series of analysis experiments to show the superiority of our method, including the sensitivity analysis, analysis of preview-based
    learning strategy and visualization.
	
	\begin{table*}[htbp]
	    \renewcommand\arraystretch{1.05}
		\caption{Top-1 accuracy~(\%)~of student networks on CIFAR-100 test set between similar network architectures. The accuracy of other methods comes from~\cite{tian2019contrastive,liu2021exploring,DBLP:conf/cvpr/ZhaoCSQL22,li2023boosting,DBLP:conf/cvpr/JinWL23} without KD. SSKD is reproduced based on author-provided code.~({\color{red}{$\downarrow n$}}) indicates the method decreases by $n$ compared with KD and ({\color{green}{$\uparrow n$}}) denotes the method exceeds $n$ compared with KD. NA denotes no result. \textbf{Bold} and \underline{underline} denote the best and the state-of-the-art results, respectively. }
		\centering
		\begin{tabular}{lccccccc}
			\toprule
			Teacher & WRN-40-2& WRN-40-2& ResNet56& ResNet110& ResNet110& ResNet32x4& VGG13\\
			Student & WRN-16-2& WRN-40-1& ResNet20& ResNet20&ResNet32& ResNet8x4& VGG8\\
			\midrule 
			Teacher& 75.61& 75.61& 72.34& 74.31& 74.31& 79.42& 74.64  \\
			Student& 73.26& 71.98& 69.06& 69.06& 71.14& 72.50& 70.36 \\
			\midrule
			KD~\cite{hinton2015distilling}& 74.92& 73.54& 70.66& 70.67& 73.08& 73.33& 72.98 \\
			FitNet~\cite{romero2014fitnets} & 73.58({\color{red}{\scriptsize{$\downarrow$}\tiny{1.34}}}) & 72.24({\color{red}{\scriptsize{$\downarrow$}\tiny{1.30}}}) & 
			69.21({\color{red}{\scriptsize{$\downarrow$}\tiny{1.45}}}) & 68.99({\color{red}{\scriptsize{$\downarrow$}\tiny{1.68}}}) & 
			71.06({\color{red}{\scriptsize{$\downarrow$}\tiny{2.02}}}) & 
			73.50({\color{green}{\scriptsize{$\uparrow$}\tiny{0.17}}}) & 
			71.02({\color{red}{\scriptsize{$\downarrow$}\tiny{1.96}}}) \\
			AT~\cite{zagoruyko2016paying} & 74.08({\color{red}{\scriptsize{$\downarrow$}\tiny{0.84}}})& 
			72.77({\color{red}{\scriptsize{$\downarrow$}\tiny{0.77}}})& 
			70.55({\color{red}{\scriptsize{$\downarrow$}\tiny{0.11}}})& 
			70.22({\color{red}{\scriptsize{$\downarrow$}\tiny{0.45}}})& 
			72.31({\color{red}{\scriptsize{$\downarrow$}\tiny{0.77}}})& 
			73.44({\color{green}{\scriptsize{$\uparrow$}\tiny{0.11}}})& 
			71.43({\color{red}{\scriptsize{$\downarrow$}\tiny{1.55}}})\\
			SP~\cite{tung2019similarity} & 73.83({\color{red}{\scriptsize{$\downarrow$}\tiny{1.09}}})& 
			72.43({\color{red}{\scriptsize{$\downarrow$}\tiny{1.11}}})& 
			69.67({\color{red}{\scriptsize{$\downarrow$}\tiny{0.99}}})& 
			70.04({\color{red}{\scriptsize{$\downarrow$}\tiny{0.63}}})& 
			72.69({\color{red}{\scriptsize{$\downarrow$}\tiny{0.39}}})& 
			72.94({\color{red}{\scriptsize{$\downarrow$}\tiny{0.39}}})& 
			72.68({\color{red}{\scriptsize{$\downarrow$}\tiny{0.20}}}) \\
			CC~\cite{peng2019correlation} & 73.56({\color{red}{\scriptsize{$\downarrow$}\tiny{1.36}}})& 
			72.21({\color{red}{\scriptsize{$\downarrow$}\tiny{1.33}}})& 
			69.63({\color{red}{\scriptsize{$\downarrow$}\tiny{1.03}}})& 
			69.48({\color{red}{\scriptsize{$\downarrow$}\tiny{1.19}}})& 
			71.48({\color{red}{\scriptsize{$\downarrow$}\tiny{1.60}}})& 
			72.97({\color{red}{\scriptsize{$\downarrow$}\tiny{0.36}}})& 
			70.71({\color{red}{\scriptsize{$\downarrow$}\tiny{2.27}}}) \\
			VID~\cite{ahn2019variational} & 74.11({\color{red}{\scriptsize{$\downarrow$}\tiny{0.81}}})& 
			73.30({\color{red}{\scriptsize{$\downarrow$}\tiny{0.24}}})& 
			70.38({\color{red}{\scriptsize{$\downarrow$}\tiny{0.28}}})& 
			70.16({\color{red}{\scriptsize{$\downarrow$}\tiny{0.51}}})& 
			72.61({\color{red}{\scriptsize{$\downarrow$}\tiny{0.47}}})& 
			73.09({\color{red}{\scriptsize{$\downarrow$}\tiny{0.24}}})& 
			71.23({\color{red}{\scriptsize{$\downarrow$}\tiny{1.75}}}) \\
			RKD~\cite{park2019relational} & 73.35({\color{red}{\scriptsize{$\downarrow$}\tiny{1.57}}})& 
			72.22({\color{red}{\scriptsize{$\downarrow$}\tiny{1.32}}})& 
			69.61({\color{red}{\scriptsize{$\downarrow$}\tiny{1.05}}})& 
			69.25({\color{red}{\scriptsize{$\downarrow$}\tiny{1.42}}})& 
			71.82({\color{red}{\scriptsize{$\downarrow$}\tiny{1.26}}})& 
			71.90({\color{red}{\scriptsize{$\downarrow$}\tiny{1.43}}})& 
			71.48({\color{red}{\scriptsize{$\downarrow$}\tiny{1.50}}}) \\
			PKT~\cite{passalis2018learning} & 74.54({\color{red}{\scriptsize{$\downarrow$}\tiny{0.38}}})& 
			73.45({\color{red}{\scriptsize{$\downarrow$}\tiny{0.09}}})& 
			70.34({\color{red}{\scriptsize{$\downarrow$}\tiny{0.32}}})& 
			70.25({\color{red}{\scriptsize{$\downarrow$}\tiny{0.42}}})& 
			72.61({\color{red}{\scriptsize{$\downarrow$}\tiny{0.47}}})& 
			73.64({\color{green}{\scriptsize{$\uparrow$}\tiny{0.31}}})& 
			72.88({\color{red}{\scriptsize{$\downarrow$}\tiny{0.10}}}) \\
			AB~\cite{heo2019knowledge}& 72.50({\color{red}{\scriptsize{$\downarrow$}\tiny{2.42}}})& 
			72.38({\color{red}{\scriptsize{$\downarrow$}\tiny{1.16}}})& 
			69.47({\color{red}{\scriptsize{$\downarrow$}\tiny{1.19}}})& 
			69.53({\color{red}{\scriptsize{$\downarrow$}\tiny{1.14}}})& 
			70.98({\color{red}{\scriptsize{$\downarrow$}\tiny{2.10}}})& 
			73.17({\color{red}{\scriptsize{$\downarrow$}\tiny{0.16}}})& 
			70.94({\color{red}{\scriptsize{$\downarrow$}\tiny{2.04}}}) \\
			FT~\cite{kim2018paraphrasing}& 73.25({\color{red}{\scriptsize{$\downarrow$}\tiny{1.67}}})& 
			71.59({\color{red}{\scriptsize{$\downarrow$}\tiny{1.95}}})& 
			69.84({\color{red}{\scriptsize{$\downarrow$}\tiny{0.82}}})& 
			70.22({\color{red}{\scriptsize{$\downarrow$}\tiny{0.45}}})& 
			72.37({\color{red}{\scriptsize{$\downarrow$}\tiny{0.71}}})& 
			72.86({\color{red}{\scriptsize{$\downarrow$}\tiny{0.47}}})& 
			70.58({\color{red}{\scriptsize{$\downarrow$}\tiny{2.40}}}) \\
			FSP~\cite{yim2017gift}& 72.91({\color{red}{\scriptsize{$\downarrow$}\tiny{2.01}}})& 
			NA& 
			69.95({\color{red}{\scriptsize{$\downarrow$}\tiny{0.71}}})& 
			70.11({\color{red}{\scriptsize{$\downarrow$}\tiny{0.56}}})& 
			71.89({\color{red}{\scriptsize{$\downarrow$}\tiny{1.19}}})& 
			72.62({\color{red}{\scriptsize{$\downarrow$}\tiny{0.71}}})& 
			70.23({\color{red}{\scriptsize{$\downarrow$}\tiny{2.75}}}) \\
			NST~\cite{huang2017like}& 73.68({\color{red}{\scriptsize{$\downarrow$}\tiny{1.24}}})& 
			72.24({\color{red}{\scriptsize{$\downarrow$}\tiny{1.30}}})& 
			69.60({\color{red}{\scriptsize{$\downarrow$}\tiny{1.06}}})& 
			69.53({\color{red}{\scriptsize{$\downarrow$}\tiny{1.14}}})& 
			71.96({\color{red}{\scriptsize{$\downarrow$}\tiny{1.12}}})& 
			73.30({\color{red}{\scriptsize{$\downarrow$}\tiny{0.03}}})& 
			71.53({\color{red}{\scriptsize{$\downarrow$}\tiny{1.45}}}) \\
			CRD~\cite{tian2019contrastive}& 75.48({\color{green}{\scriptsize{$\uparrow$}\tiny{0.56}}})& 
			74.14({\color{green}{\scriptsize{$\uparrow$}\tiny{0.60}}})& 
			71.16({\color{green}{\scriptsize{$\uparrow$}\tiny{0.50}}})& 
			71.46({\color{green}{\scriptsize{$\uparrow$}\tiny{0.79}}})& 
			73.48({\color{green}{\scriptsize{$\uparrow$}\tiny{0.40}}})& 
			75.51({\color{green}{\scriptsize{$\uparrow$}\tiny{2.18}}})& 
			73.94({\color{green}{\scriptsize{$\uparrow$}\tiny{0.96}}}) \\
			
			SSKD~\cite{xu2020knowledge}& 74.87({\color{red}{\scriptsize{$\downarrow$}\tiny{0.05}}})&
			74.59({\color{green}{\scriptsize{$\uparrow$}\tiny{1.05}}})&
			70.20({\color{red}{\scriptsize{$\downarrow$}\tiny{0.46}}})&
			69.85({\color{red}{\scriptsize{$\downarrow$}\tiny{0.82}}})&
			72.63({\color{red}{\scriptsize{$\downarrow$}\tiny{0.45}}})&
			75.69({\color{green}{\scriptsize{$\uparrow$}\tiny{2.36}}})&
			74.35({\color{green}{\scriptsize{$\uparrow$}\tiny{1.37}}})
			\\
			ICKD~\cite{liu2021exploring}& 75.64({\color{green}{\scriptsize{$\uparrow$}\tiny{0.72}}})& 
			74.33({\color{green}{\scriptsize{$\uparrow$}\tiny{0.79}}})& 
			71.76({\color{green}{\scriptsize{$\uparrow$}\tiny{1.10}}})& 
			71.68({\color{green}{\scriptsize{$\uparrow$}\tiny{1.01}}})& 
			73.89({\color{green}{\scriptsize{$\uparrow$}\tiny{0.81}}})& 
			75.25({\color{green}{\scriptsize{$\uparrow$}\tiny{1.92}}})& 
			73.42({\color{green}{\scriptsize{$\uparrow$}\tiny{0.44}}}) \\
            LSKD~\cite{li2023boosting}& 74.10({\color{red}{\scriptsize{$\downarrow$}\tiny{0.82}}})& 
			75.22({\color{green}{\scriptsize{$\uparrow$}\tiny{1.68}}})& 
			71.43({\color{green}{\scriptsize{$\uparrow$}\tiny{0.77}}})& 
			71.38({\color{green}{\scriptsize{$\uparrow$}\tiny{0.71}}})& 
            NA& 
			74.38({\color{green}{\scriptsize{$\uparrow$}\tiny{1.05}}})& 
			NA \\
            ReviewKD~\cite{DBLP:conf/cvpr/Chen0ZJ21}& 76.12({\color{green}{\scriptsize{$\uparrow$}\tiny{1.20}}})& 
			75.09({\color{green}{\scriptsize{$\uparrow$}\tiny{1.55}}})& 
			71.89({\color{green}{\scriptsize{$\uparrow$}\tiny{1.23}}})& 
			NA& 
			73.89({\color{green}{\scriptsize{$\uparrow$}\tiny{0.81}}})& 
			75.63({\color{green}{\scriptsize{$\uparrow$}\tiny{2.30}}})& 
			74.84({\color{green}{\scriptsize{$\uparrow$}\tiny{1.86}}}) \\
            DKD~\cite{DBLP:conf/cvpr/ZhaoCSQL22}& 76.24({\color{green}{\scriptsize{$\uparrow$}\tiny{1.32}}})& 
			74.81({\color{green}{\scriptsize{$\uparrow$}\tiny{1.27}}})& 
			71.97({\color{green}{\scriptsize{$\uparrow$}\tiny{1.31}}})& 
			\underline{71.73}({\color{green}{\scriptsize{$\uparrow$}\tiny{1.06}}})& 
			\underline{74.11}({\color{green}{\scriptsize{$\uparrow$}\tiny{1.03}}})& 
			76.32({\color{green}{\scriptsize{$\uparrow$}\tiny{2.99}}})& 
			74.68({\color{green}{\scriptsize{$\uparrow$}\tiny{1.70}}}) \\
            MLD~\cite{DBLP:conf/cvpr/JinWL23}&  \underline{76.63}({\color{green}{\scriptsize{$\uparrow$}\tiny{1.71}}}) &
			\underline{75.35}({\color{green}{\scriptsize{$\uparrow$}\tiny{1.81}}})& 
			\underline{72.19}({\color{green}{\scriptsize{$\uparrow$}\tiny{1.53}}})& 
			NA& 
			\underline{74.11}({\color{green}{\scriptsize{$\uparrow$}\tiny{1.03}}})& 
			\underline{77.08}({\color{green}{\scriptsize{$\uparrow$}\tiny{3.75}}})& 
			\underline{75.18}({\color{green}{\scriptsize{$\uparrow$}\tiny{2.20}}}) \\

            \midrule
			PCKD (w/o $\mathcal{L}_{KD}$)&76.52({\color{green}{\scriptsize{$\uparrow$}\tiny{1.60}}})&
			76.22({\color{green}{\scriptsize{$\uparrow$}\tiny{2.68}}})&
			71.92({\color{green}{\scriptsize{$\uparrow$}\tiny{1.26}}})&
			72.05({\color{green}{\scriptsize{$\uparrow$}\tiny{1.38}}})&
			74.33({\color{green}{\scriptsize{$\uparrow$}\tiny{1.25}}})& 
            76.89({\color{green}{\scriptsize{$\uparrow$}\tiny{3.56}}})&
			75.61({\color{green}{\scriptsize{$\uparrow$}\tiny{2.63}}})\\
			PCKD (Ours)&
			\textbf{76.97}({\color{green}{\scriptsize{$\uparrow$}\tiny{2.05}}})&
			\textbf{76.24}({\color{green}{\scriptsize{$\uparrow$}\tiny{2.70}}})&
			\textbf{72.21}({\color{green}{\scriptsize{$\uparrow$}\tiny{1.55}}})&
			\textbf{72.55}({\color{green}{\scriptsize{$\uparrow$}\tiny{1.88}}})&
			\textbf{75.12}({\color{green}{\scriptsize{$\uparrow$}\tiny{2.04}}})&
			\textbf{77.14}({\color{green}{\scriptsize{$\uparrow$}\tiny{3.81}}})&
			\textbf{75.80}({\color{green}{\scriptsize{$\uparrow$}\tiny{2.82}}})\\
			\bottomrule
		\end{tabular}
		\label{t1}
	\end{table*}

	\subsection{Experimental Settings}\label{Experimental Settings}
	\emph{1) Datasets:} We adopted four commonly-used challenging datasets for evaluation, including CIFAR-100~\cite{krizhevsky2009learning}, ImageNet~\cite{deng2009imagenet}, TinyImageNet~\cite{deng2009imagenet}, and STL-10~\cite{coates2011analysis}.

    \begin{itemize}
	\item CIFAR-100~\cite{krizhevsky2009learning}  contains $50K$ training images and $10K$ validation images that belong to 100 categories, where each RGB image is of size $32 \times 32$.
	
	\item ImageNet~\cite{deng2009imagenet} has $1.2$ million training images and $50K$ test images of $1,000$ categories. 
	Each image is resized to $224 \times 224 \times 3$ as the input to the convolutional neural network.
	
	\item TinyImageNet~\cite{deng2009imagenet} is a subset of ImageNet with fewer categories and smaller image size. It includes $200$ classes, each with $500$, $50$, and $50$ images for training, validation, and test, respectively.
	Each image is downsized to $64 \times 64$.
	
	\item STL-10~\cite{coates2011analysis} consists of $13,000$ images with $10$ classes, among which $5K$ images are partitioned for training while the remaining $8K$ images for testing. All the images are RGB images with size of $96 \times 96$.
	
	\end{itemize}
 
	\textbf{Compared Methods}. We mainly compared the performance of our method with the following state-of-the-art methods.
	
	\begin{itemize}
	
	\item {KD~\cite{hinton2015distilling}} transfers the predictions of the network as knowledge.
	
	\item {CRD~\cite{tian2019contrastive}} maximizes a lower-bound to the mutual information between the teacher and student representation to distill the structural knowledge.
	
	\item {SSKD~\cite{xu2020knowledge}} exploits self-supervised contrastive prediction as an auxiliary task for mining structural knowledge.
	
	\item {ICKD~\cite{liu2021exploring}} mines structural information of inter-channel correlation as knowledge.

    \item {ReviewKD~\cite{DBLP:conf/cvpr/Chen0ZJ21}} explores the connection path cross levels between teacher and student networks.

    \item {DKD~\cite{DBLP:conf/cvpr/ZhaoCSQL22}} enhances the teachers’ distilling performance via decoupling standard KD.

    \item {LSKD~\cite{li2023boosting}} studies the potential capabilities of KD and label smoothing in sequential knowledge transfer.

    \item {MLD~\cite{DBLP:conf/cvpr/JinWL23}} mines stronger logits knowledge through multi-level prediction alignment.

	\end{itemize}
	
	\emph{3) Implementation Details}: On the CIFAR-100 dataset, we adopted a similar setting as~\cite{tian2019contrastive} for a fair comparison.
	Specifically, images are horizontally flipped with a probability of $0.5$ and normalized by the mean and standard deviation. We expanded the samples by rotating an image by $90^{\circ}$ in turn into four images. We empirically set the weight factors in Eq.~(\ref{e8}) as $\alpha=1$,  $\beta_{ca}=1$, and $\beta_{fa}=20$. $\beta_{cc}$ is set to $0.02$ for the student network ResNet20 and $0.05$ for other student networks.
	For heterogeneous teacher and student backbones, the dimension of category centers corresponding to the weights of the fully connected layers might be different, where  we set $\beta_{ca}=0$.
	The SGD optimizer~\cite{sutskever2013importance} is used to train the student model with batch size $64$, momentum $0.9$, and weight decay $5e-4$. The learning rate starts from $0.05$ and is multiplied by $0.1$ at $150$, $180$, and $210$ epochs with $240$ epochs in total.
	
	On the ImageNet dataset, each cropped image is resized to $224\times224$ and flipped with a probability of $0.5$. We normalized three channels with the mean and standard deviation and expanded the samples by rotating an image by $90^{\circ}$ in turn into four images. We used the SGD optimizer to train the student model for $120$ epochs with a total batch size $1024$, where we used 16 GPUs and each with batch size $64$, momentum $0.9$, weight decay $1e-4$, and warm-up epoch $5$. The initial learning rate is $0.4$ and decayed by $0.1$ at $40$, $70$, $100$, and $110$ epochs. The weight factors in Eq.~(\ref{e8}) are $\alpha=1$, $\beta_{cc}=0.005$, $\beta_{ca}=1$, $\beta_{fa}=20$.

	On STL-10 and TinyImageNet, similar to~\cite{tian2019contrastive}, we first trained the student model with the guidance of the teacher model on the CIFAR-100 dataset.
	Then we froze the feature extraction layer, then trained the fully connected layer on the STL-10 and TinyImageNet datasets, after which we tested on these two datasets.
	ResNet110 and ResNet20~\cite{he2016deep} are selected as teacher and student models, respectively. We used the SGD optimizer to train the student model for $100$ epochs with a batch size $64$, momentum $0.9$, and weight decay $1e-4$. The initial learning rate is $0.1$ and decayed by $0.1$ at $30$, $60$, and $90$ epochs.
	
	\begin{table*}[htbp]
	    \renewcommand\arraystretch{1.05}
		\caption{Top-1 accuracy~(\%)~of student networks on CIFAR-100 test set across different network architectures. The accuracy of other methods comes from~\cite{tian2019contrastive,liu2021exploring,DBLP:conf/cvpr/ZhaoCSQL22,li2023boosting,DBLP:conf/cvpr/JinWL23} without KD. SSKD and ICKD are reproduced by us based on the author-provided code. ~({\color{red}{$\downarrow n$}})~indicates the method decreases by $n$ compared with KD and~({\color{green}{$\uparrow n$}})~denotes the method exceeds $n$ compared with KD. NA denotes no result. \textbf{Bold} and \underline{underline} denote the best and the state-of-the-art results, respectively.}
		\centering
		\begin{tabular}{lcccccc}
			\toprule
			Teacher & VGG13& ResNet50& ResNet50& ResNet32x4& ResNet32x4& WRN-40-2\\
			Student & MobileNetV2& MobileNetV2& VGG8& ShuffleNetV1& ShuffleNetV2& ShuffleNetV1 \\
			\midrule 
			Teacher& 74.64& 79.34& 79.34& 79.42& 79.42& 75.61 \\
			Student& 64.60& 64.60& 70.36& 70.50& 71.82& 70.50 \\
			\midrule
			KD~\cite{hinton2015distilling}& 67.37& 67.35& 73.81& 74.07& 74.45& 74.83 \\
			FitNet~\cite{romero2014fitnets}& 64.14({\color{red}{\scriptsize{$\downarrow$}\tiny{3.23}}}) & 63.16({\color{red}{\scriptsize{$\downarrow$}\tiny{4.19}}}) & 
			70.69({\color{red}{\scriptsize{$\downarrow$}\tiny{3.12}}}) & 73.59({\color{red}{\scriptsize{$\downarrow$}\tiny{0.48}}}) & 
			73.54({\color{red}{\scriptsize{$\downarrow$}\tiny{0.91}}}) & 
			73.73({\color{red}{\scriptsize{$\downarrow$}\tiny{1.10}}}) \\
			AT~\cite{zagoruyko2016paying}& 59.40({\color{red}{\scriptsize{$\downarrow$}\tiny{7.97}}})& 
			58.58({\color{red}{\scriptsize{$\downarrow$}\tiny{8.77}}})& 
			71.84({\color{red}{\scriptsize{$\downarrow$}\tiny{1.97}}})& 
			71.73({\color{red}{\scriptsize{$\downarrow$}\tiny{2.34}}})& 
			72.73({\color{red}{\scriptsize{$\downarrow$}\tiny{1.72}}})& 
			73.32({\color{red}{\scriptsize{$\downarrow$}\tiny{1.51}}})  \\
			SP~\cite{tung2019similarity} & 66.30({\color{red}{\scriptsize{$\downarrow$}\tiny{1.07}}})& 
			68.08({\color{green}{\scriptsize{$\uparrow$}\tiny{0.73}}})& 
			73.34({\color{red}{\scriptsize{$\downarrow$}\tiny{0.47}}})& 
			73.48({\color{red}{\scriptsize{$\downarrow$}\tiny{0.59}}})& 
			74.56({\color{green}{\scriptsize{$\uparrow$}\tiny{0.11}}})& 
			74.52({\color{red}{\scriptsize{$\downarrow$}\tiny{0.31}}}) \\
			CC~\cite{peng2019correlation} & 64.86({\color{red}{\scriptsize{$\downarrow$}\tiny{2.51}}})& 
			65.43({\color{red}{\scriptsize{$\downarrow$}\tiny{1.92}}})& 
			70.25({\color{red}{\scriptsize{$\downarrow$}\tiny{3.56}}})& 
			71.14({\color{red}{\scriptsize{$\downarrow$}\tiny{2.93}}})& 
			71.29({\color{red}{\scriptsize{$\downarrow$}\tiny{3.16}}})& 
			71.38({\color{red}{\scriptsize{$\downarrow$}\tiny{3.45}}})\\
			VID~\cite{ahn2019variational} & 65.56({\color{red}{\scriptsize{$\downarrow$}\tiny{1.81}}})& 
			67.57({\color{green}{\scriptsize{$\uparrow$}\tiny{0.22}}})& 
			70.30({\color{red}{\scriptsize{$\downarrow$}\tiny{3.51}}})& 
			73.38({\color{red}{\scriptsize{$\downarrow$}\tiny{0.69}}})& 
			73.40({\color{red}{\scriptsize{$\downarrow$}\tiny{1.05}}})& 
			73.61({\color{red}{\scriptsize{$\downarrow$}\tiny{1.22}}}) \\
			RKD~\cite{park2019relational} & 64.52({\color{red}{\scriptsize{$\downarrow$}\tiny{2.85}}})& 
			64.43({\color{red}{\scriptsize{$\downarrow$}\tiny{2.92}}})& 
			71.50({\color{red}{\scriptsize{$\downarrow$}\tiny{2.31}}})& 
			72.28({\color{red}{\scriptsize{$\downarrow$}\tiny{1.79}}})& 
			73.21({\color{red}{\scriptsize{$\downarrow$}\tiny{1.24}}})& 
			72.21({\color{red}{\scriptsize{$\downarrow$}\tiny{2.62}}}) \\
			PKT~\cite{passalis2018learning}& 67.13({\color{red}{\scriptsize{$\downarrow$}\tiny{0.24}}})& 
			66.52({\color{red}{\scriptsize{$\downarrow$}\tiny{0.83}}})& 
			73.01({\color{red}{\scriptsize{$\downarrow$}\tiny{0.80}}})& 
			74.10({\color{green}{\scriptsize{$\uparrow$}\tiny{0.03}}})& 
			74.69({\color{green}{\scriptsize{$\uparrow$}\tiny{0.24}}})& 
			73.89({\color{red}{\scriptsize{$\downarrow$}\tiny{0.94}}}) \\
			AB~\cite{heo2019knowledge}& 66.06({\color{red}{\scriptsize{$\downarrow$}\tiny{1.31}}})& 
			67.20({\color{red}{\scriptsize{$\downarrow$}\tiny{0.15}}})& 
			70.65({\color{red}{\scriptsize{$\downarrow$}\tiny{3.16}}})& 
			73.55({\color{red}{\scriptsize{$\downarrow$}\tiny{0.52}}})& 
			74.31({\color{red}{\scriptsize{$\downarrow$}\tiny{0.14}}})& 
			73.34({\color{red}{\scriptsize{$\downarrow$}\tiny{1.49}}}) \\
			FT~\cite{kim2018paraphrasing}& 61.78({\color{red}{\scriptsize{$\downarrow$}\tiny{5.59}}})& 
			60.99({\color{red}{\scriptsize{$\downarrow$}\tiny{6.36}}})& 
			70.29({\color{red}{\scriptsize{$\downarrow$}\tiny{3.52}}})& 
			71.75({\color{red}{\scriptsize{$\downarrow$}\tiny{2.32}}})& 
			72.50({\color{red}{\scriptsize{$\downarrow$}\tiny{1.95}}})& 
			72.03({\color{red}{\scriptsize{$\downarrow$}\tiny{2.80}}}) \\
			NST~\cite{huang2017like}& 58.16({\color{red}{\scriptsize{$\downarrow$}\tiny{9.21}}})& 
			64.96({\color{red}{\scriptsize{$\downarrow$}\tiny{2.39}}})& 
			71.28{\color{red}{\scriptsize{$\downarrow$}\tiny{2.53}}})& 
			74.12({\color{green}{\scriptsize{$\uparrow$}\tiny{0.05}}})& 
			74.68({\color{green}{\scriptsize{$\uparrow$}\tiny{0.23}}})& 
			74.89({\color{green}{\scriptsize{$\uparrow$}\tiny{0.06}}})\\
			CRD~\cite{tian2019contrastive}& 69.73({\color{green}{\scriptsize{$\uparrow$}\tiny{2.36}}})& 
			69.11({\color{green}{\scriptsize{$\uparrow$}\tiny{1.76}}})& 
			74.30({\color{green}{\scriptsize{$\uparrow$}\tiny{0.49}}})& 
			75.11({\color{green}{\scriptsize{$\uparrow$}\tiny{1.04}}})& 
			75.65({\color{green}{\scriptsize{$\uparrow$}\tiny{1.20}}})& 
			76.05({\color{green}{\scriptsize{$\uparrow$}\tiny{1.22}}}) \\
			SSKD~\cite{xu2020knowledge}& 70.43({\color{green}{\scriptsize{$\uparrow$}\tiny{3.06}}})& 
			\underline{71.04}({\color{green}{\scriptsize{$\uparrow$}\tiny{3.69}}})&
			\underline{74.87}({\color{green}{\scriptsize{$\uparrow$}\tiny{1.06}}})&
			\underline{77.58}({\color{green}{\scriptsize{$\uparrow$}\tiny{3.51}}})&
			77.88({\color{green}{\scriptsize{$\uparrow$}\tiny{3.43}}})&
			76.59({\color{green}{\scriptsize{$\uparrow$}\tiny{1.76}}})
			\\
			ICKD~\cite{liu2021exploring}& 67.19({\color{red}{\scriptsize{$\downarrow$}\tiny{0.18}}})& 
			68.00({\color{green}{\scriptsize{$\uparrow$}\tiny{0.65}}})& 
			73.24({\color{red}{\scriptsize{$\downarrow$}\tiny{0.57}}})& 
			73.93({\color{red}{\scriptsize{$\downarrow$}\tiny{0.14}}})& 
			75.47({\color{green}{\scriptsize{$\uparrow$}\tiny{1.02}}})& 
			74.78({\color{red}{\scriptsize{$\downarrow$}\tiny{0.05}}}) \\
            LSKD~\cite{li2023boosting}& NA& 
			NA& 
			NA& 
			NA& 
			75.13({\color{green}{\scriptsize{$\uparrow$}\tiny{0.68}}})& 
			76.33({\color{green}{\scriptsize{$\uparrow$}\tiny{1.50}}}) \\ 
            ReviewKD~\cite{DBLP:conf/cvpr/Chen0ZJ21}& 70.37({\color{green}{\scriptsize{$\uparrow$}\tiny{3.00}}})& 
			69.89({\color{green}{\scriptsize{$\uparrow$}\tiny{2.54}}})& 
			NA& 
			77.45({\color{green}{\scriptsize{$\uparrow$}\tiny{3.38}}})& 
			77.78({\color{green}{\scriptsize{$\uparrow$}\tiny{3.33}}})& 
			77.14({\color{green}{\scriptsize{$\uparrow$}\tiny{2.31}}}) \\
            DKD~\cite{DBLP:conf/cvpr/ZhaoCSQL22}& 69.71({\color{green}{\scriptsize{$\uparrow$}\tiny{2.34}}})& 
			70.35({\color{green}{\scriptsize{$\uparrow$}\tiny{3.00}}})&  
			NA& 
			76.45({\color{green}{\scriptsize{$\uparrow$}\tiny{2.38}}})& 
			77.07({\color{green}{\scriptsize{$\uparrow$}\tiny{2.62}}})& 
			76.70({\color{green}{\scriptsize{$\uparrow$}\tiny{1.87}}}) \\
            MLD~\cite{DBLP:conf/cvpr/JinWL23}&  \underline{70.57}({\color{green}{\scriptsize{$\uparrow$}\tiny{3.20}}}) &
			\underline{71.04}({\color{green}{\scriptsize{$\uparrow$}\tiny{3.69}}})& 
			NA& 
			77.18({\color{green}{\scriptsize{$\uparrow$}\tiny{3.11}}}) & 
			\underline{78.44}({\color{green}{\scriptsize{$\uparrow$}\tiny{3.99}}})& 
			\underline{77.44}({\color{green}{\scriptsize{$\uparrow$}\tiny{2.61}}}) \\
   
            \midrule
			PCKD (w/o $\mathcal{L}_{KD}$)&72.18({\color{green}{\scriptsize{$\uparrow$}\tiny{4.81}}})& 
			\textbf{72.12}({\color{green}{\scriptsize{$\uparrow$}\tiny{4.77}}})&
			76.27({\color{green}{\scriptsize{$\uparrow$}\tiny{2.46}}})&
			78.90({\color{green}{\scriptsize{$\uparrow$}\tiny{4.83}}})&
			\textbf{79.91}({\color{green}{\scriptsize{$\uparrow$}\tiny{5.46}}})&
			78.22({\color{green}{\scriptsize{$\uparrow$}\tiny{3.39}}})\\
			PCKD (Ours)&
			\textbf{72.37}({\color{green}{\scriptsize{$\uparrow$}\tiny{5.00}}})&
			71.85({\color{green}{\scriptsize{$\uparrow$}\tiny{4.50}}})&
			\textbf{76.32}({\color{green}{\scriptsize{$\uparrow$}\tiny{2.51}}})&
			\textbf{79.21}({\color{green}{\scriptsize{$\uparrow$}\tiny{5.14}}})&
			79.61({\color{green}{\scriptsize{$\uparrow$}\tiny{5.16}}})&
			\textbf{78.44}({\color{green}{\scriptsize{$\uparrow$}\tiny{3.61}}})\\
			\bottomrule
		\end{tabular}
		\label{t2}
	\end{table*}
	
	\begin{table*}[htbp]
	    \renewcommand\arraystretch{1.05}
		\caption{Top-1 and Top-5 accuracy~(\%)~of student network on ImageNet validation set. We use ResNet34 as the teacher model and ResNet18 as the student model~\cite{he2016deep}. Results of other methods are from~\cite{tian2019contrastive,xu2020knowledge,DBLP:conf/cvpr/JinWL23}. }
		%\captionsetup[singlelinecheck=false]{}
		\centering
		\begin{tabular}{c|ccccccccc|c}
			\toprule
			Accuracy &  Student& KD~\cite{hinton2015distilling}& AT~\cite{zagoruyko2016paying}& CRD~\cite{tian2019contrastive}& SSKD~\cite{xu2020knowledge}& 
            ReviewKD~\cite{DBLP:conf/cvpr/Chen0ZJ21} &
            DKD~\cite{DBLP:conf/cvpr/ZhaoCSQL22} &
            MLD~\cite{DBLP:conf/cvpr/JinWL23} 
            & Ours &Teacher\\
			\midrule
			Top-1 & 69.75& 70.66& 70.70& 71.38& 71.62&  71.61 & 71.70 & \underline{71.90} &\textbf{72.01} &  73.31 \\
			\midrule 
			Top-5& 89.07& 89.88& 90.00& 90.49& \underline{90.67}& 90.51  &  90.41 & 90.55 &\textbf{90.69}  &91.42 \\
			\bottomrule
		\end{tabular}
		\label{t3}
	\end{table*}
	
	\begin{table*}[htbp] 
	    \renewcommand\arraystretch{1.05}
		\caption{Top-1 accuracy~(\%) of transfer learning based on the architecture combination of ResNet110 and ResNet20~\cite{he2016deep}. The results of these methods are without KD.
		We reproduced other methods according to~\cite{tian2019contrastive,liu2021exploring,xu2020knowledge}.}
		\centering
		\begin{tabular}{l|cccccccc|c}
			\toprule
			Transferred Dataset& Student& KD~\cite{hinton2015distilling}& FitNet~\cite{romero2014fitnets}& AT~\cite{zagoruyko2016paying}& CRD~\cite{tian2019contrastive}& ICKD~\cite{liu2021exploring}& SSKD~\cite{xu2020knowledge}&PCKD&Teacher\\
			\midrule
			CIFAR-100$\rightarrow$STL-10&63.15 & 62.96& 64.10& 63.75& 65.30& 65.51& \underline{67.11}& \textbf{68.85}& 64.91  \\
			\midrule 
			CIFAR-100$\rightarrow$TinyImageNet& 28.69& 29.21& 28.28& 28.36& 31.78& 29.80& \underline{31.99}& \textbf{34.88}&29.54 \\
			\bottomrule
		\end{tabular}
		\label{t4}
	\end{table*}
	
	\subsection{Comparison with the State-of-the-art Methods}
	
	\noindent\textbf{Results on CIFAR-100}. To comprehensively evaluate the performance, similar to~\cite{tian2019contrastive},
	we compared the results under many different network structures, such as ResNet~\cite{he2016deep}, WideResNet~\cite{zagoruyko2016wide}, VGGNet~\cite{simonyan2014very}, MobileNet~\cite{howard2017mobilenets}, ShuffleNet~\cite{zhang2018shufflenet}, and its variants~\cite{sandler2018mobilenetv2,ma2018shufflenet,xie2017aggregated}.
	We conducted two sets of experiments, including the same type of network structure and different types of network structures.
	The results of the same teacher and student architecture are shown in Table~\ref{t1}.
	It is obvious that our method outperforms all other methods by a large margin under all these seven architecture combinations.
	Specifically, the absolute average improvement of our PCKD method without KD loss over the vanilla KD is $2.05\%$, while the second best approach MLD~\cite{DBLP:conf/cvpr/JinWL23} improves KD by an average of $2.005\%$.
	By incorporating the KD loss, PCKD can improve the average results by $2.41\%$.
	Moreover, we also observed that under several architectures, such as WRN-40-2 and WRN-16-2, VGG13 and VGG8, the student model learned by our method achieves better performance than the teacher model, which is very challenging.
	The above results can well demonstrate the superiority of our PCKD method over existing state-of-the-art methods.
 
	For different styles of teacher and student structures, the results are shown in Table~\ref{t2}.
	We can also see that PCKD consistently outperforms all other methods under all different architectures, and the improvement is more significant than that of the same architecture style.
	On these six settings, PCKD without KD loss absolutely improves vanilla KD by an average of $4.29\%$. Among these related methods, MLD~\cite{DBLP:conf/cvpr/JinWL23} achieves the second best results, which improves the KD by $3.32\%$, but it is still $0.97\%$ lower than our method.
	Interestingly, the performance of ICKD is lower than the vanilla KD in most cases, which demonstrates that it mainly works well under the same architecture style.
	In comparison, our method is very robust and achieves significantly superior results than all these methods under various settings, which can be largely attributed to the discriminative feature learning ability of our contrastive learning strategy and preview strategy.
    
	Based on the above results, we also found that logit based KL-divergence loss is very effective for other response, feature, and relation-based KD methods, such as FitNets~\cite{romero2014fitnets}, AT~\cite{zagoruyko2016paying}, RKD~\cite{park2019relational}, and FSP~\cite{yim2017gift}, since their results are even lower than the vanilla KD without this loss.
	In contrast, these contrastive learning based methods, such as CRD, SSKD, and PCKD, do not rely on this constraint. The main reason might be that category information is well incorporated during the construction of positive and negative pairs.
	Moreover, compared with CRD, our improvement is also very significant.

    \hspace*{\fill}

	\noindent \textbf{Results on ImageNet}. Top-1 and Top-5 accuracies of image classification on ImageNet are reported in Table~\ref{t3}, 
	PCKD also achieves very good results which can demonstrate the effectiveness and scalability of our knowledge distillation method on the large-scale and high-resolution dataset.
    It’s worth mentioning that the performance of PCKD is better than the most state-of-the-art results of logits distillation methods.
    Besides, our method achieves better accuracy than other contrastive learning methods, showing that our method is very effective in exploiting structural knowledge.

	\subsection{Transferability of representations}
	
    We also evaluated the representation transferability ability of different knowledge distillation methods. The results are presented in Table~\ref{t4}. 
	It shows that our approach obviously surpasses other knowledge distillation methods on two transferred datasets. Besides, our method outperforms CRD~\cite{tian2019contrastive} by a large margin, demonstrating the effectiveness of transferring the category representation, and the relation between feature and category centers.
    This experiment strongly validates the effectiveness of our proposed PCKD. 
	
    \begin{table*}[htbp] 
   
	\renewcommand\arraystretch{1.05}
        \footnotesize 
		\caption{Ablation study of PCKD on CIFAR-100.~({\color{green}{$\uparrow n$}})~denotes the method exceeds $n$ compared with $\mathcal{L}_{KD}$. }\centering 
		\begin{tabular}{lcccccc}
			\toprule
			\multirow{2}{*}{Methods} & \multirow{2}{*}{KD} & \multirow{2}{*}{CKD} & \multirow{2}{*}{Preview}&  WRN-40-2 & ResNet110& WRN-40-2\\
			&  & &  & WRN-16-2 & ResNet20& ShuffleNetV1\\
			\midrule
			$\mathcal{L}_{KD}$ & $\checkmark$ &   &   & 74.84& 70.76& 75.37 \\
			
			$\mathcal{L}_{KD}+Preview$ & $\checkmark$ &   & $\checkmark$ & 75.08({\color{green}{\scriptsize{$\uparrow$}\tiny{0.24}}})& 71.02({\color{green}{\scriptsize{$\uparrow$}\tiny{0.26}}}) & 75.48({\color{green}{\scriptsize{$\uparrow$}\tiny{0.11}}}) \\
			
			$\mathcal{L}_{CKD}$ & & $\checkmark$ &   & 76.31({\color{green}{\scriptsize{$\uparrow$}\tiny{1.47}}})& 72.04({\color{green}{\scriptsize{$\uparrow$}\tiny{1.28}}})& 78.10({\color{green}{\scriptsize{$\uparrow$}\tiny{2.73}}}) \\
			$\mathcal{L}_{KD}+\mathcal{L}_{CKD}$ & $\checkmark$ & $\checkmark$ &  & 76.42({\color{green}{\scriptsize{$\uparrow$}\tiny{1.58}}})& 72.15({\color{green}{\scriptsize{$\uparrow$}\tiny{1.39}}})& 78.15({\color{green}{\scriptsize{$\uparrow$}\tiny{2.78}}}) \\
			$\mathcal{L}_{PCKD}$ (w/o $\mathcal{L}_{KD}$) & & $\checkmark$ & $\checkmark$ & 76.52({\color{green}{\scriptsize{$\uparrow$}\tiny{1.68}}})& 72.05({\color{green}{\scriptsize{$\uparrow$}\tiny{1.29}}})& 78.22({\color{green}{\scriptsize{$\uparrow$}\tiny{2.85}}}) \\
			 $\mathcal{L}_{PCKD}$ & $\checkmark$ & $\checkmark$ & $\checkmark$ & \textbf{76.97}({\color{green}{\scriptsize{$\uparrow$}\tiny{2.13}}})& \textbf{72.55}({\color{green}{\scriptsize{$\uparrow$}\tiny{1.79}}})& \textbf{78.44}({\color{green}{\scriptsize{$\uparrow$}\tiny{3.07}}}) \\
			\bottomrule
		\end{tabular}
		\label{tab:ablation}
	\end{table*}
	
	\subsection{Ablation Study} \label{Ablation Study}
	In order to verify the effectiveness of each module in our proposed method, we selected two architecture combinations of the same style (WRN-40-2 to WRN-16-2~\cite{zagoruyko2016wide} and ResNet110 to ResNet20~\cite{he2016deep}) and one architecture combination of a different style~(WRN-40-2 to ShuffleNetV1~\cite{zhang2018shufflenet}) to perform the ablation study on CIFAR-100.
	
	The results are shown in Table~\ref{tab:ablation}, where $\mathcal{L}_{CKD}$ denotes $\mathcal{L}_{PCKD}$ without KD loss and preview strategy. According to the results of lines 1 and 3, the accuracy of CKD significantly exceeds KD, indicating CKD is superior to KD. By comparing the results of the first two lines, lines 3 and
    5, lines 4 and 6, we can see that the performance can be consistently improved by adding our preview strategy. Besides, $\mathcal{L}_{KD}+\mathcal{L}_{CKD}$ obtains much better results than $\mathcal{L}_{KD}$, and $\mathcal{L}_{PCKD}$ also realizes better results than the case without contrastive learning strategy, demonstrating the effectiveness of our category-based contrastive learning. 

    To demonstrate the effectiveness of each module of~CKD. We conducted ablation studies on CIFAR-100 dataset using two network combinations (WRN-40-2 to WRN-16-2 and ResNet110 to ResNet20). As shown in Table~\ref{ckd}, $\mathcal{L}_{FA}$ keeps the features of students and teachers aligned, and Top-1 accuracy is significantly improved by $2.44\%$ and $2.47\%$ respectively. With the addition of $\mathcal{L}_{CA}$ and $\mathcal{L}_{CC}$, Top-1 accuracy is further improved to the highest level. However, when $\mathcal{L}_{FA}$ is removed, $\mathcal{L}_{CA}$ and $\mathcal{L}_{CC}$ improve little and may even be counterproductive.
    This occurs because when features of the student network and the teacher network are inconsistent, the knowledge of discriminative category centers cannot be further mined, which demonstrates the robustness and generalization ability of our contributions.

    \begin{table}
      \centering
      \caption{Ablation study of CKD on CIFAR-100.}
      \label{ablation_module}
      
      \setlength{\tabcolsep}{2mm}{
      \begin{tabular}{cccccc}
        \toprule
          \multirow{2}{*}{$\mathcal{L}_{FA}$} & \multirow{2}{*}{$\mathcal{L}_{CA}$} & \multirow{2}{*}{$\mathcal{L}_{CC}$} & WRN-40-2  &  ResNet110 \\
			&  & & WRN-16-2  & ResNet20 \\
        \midrule
        $\times$ & $\times$ & $\times$ & 73.71
 & 69.35 \\
        $\checkmark$ & $\times$ & $\times$ &  76.15 & 71.82 \\
        $\times$ & $\checkmark$ &$\times$ &  73.24 & 69.31 \\
         $\times$ & $\times$  & $\checkmark$& 73.76 & 69.30 \\
         $\times$ & $\checkmark$ &  $\checkmark$& 73.78 & 69.38 \\
        $\checkmark$ & $\checkmark$ & $\times$ & 76.20 & 71.88 \\
        $\checkmark$ & $\times$ & $\checkmark$ & 76.26 &  72.01\\
        $\checkmark$ & $\checkmark$ &  $\checkmark$ & \textbf{76.31} & \textbf{72.04} \\
      \bottomrule
    \end{tabular}}
    \label{ckd}
    \end{table}

    \begin{figure*}[!t]
		\centering
		\includegraphics[width=7.1in]{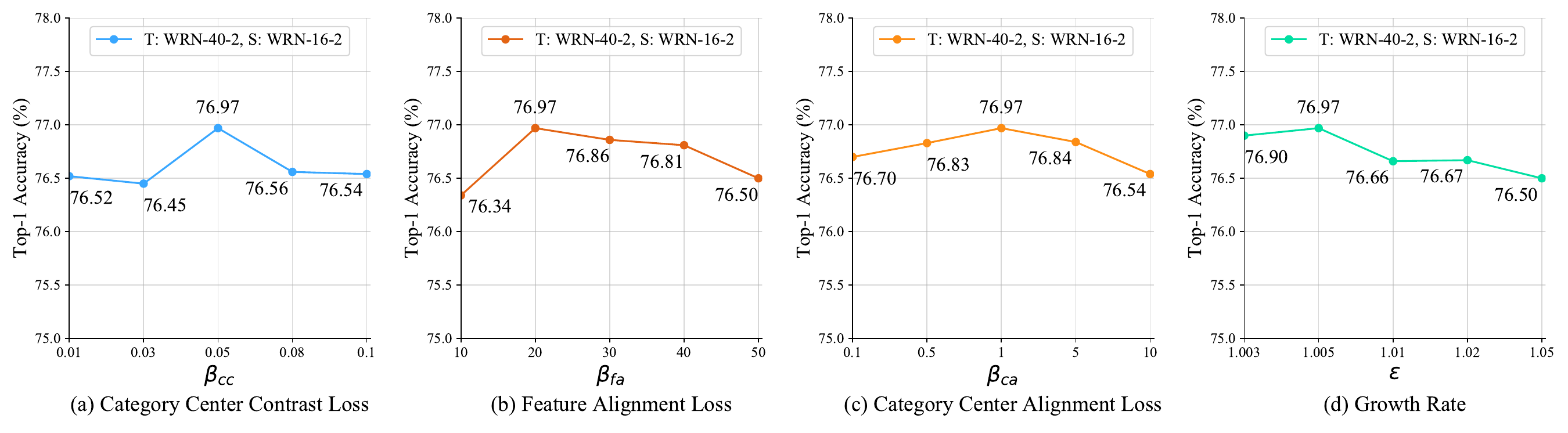}
		\caption{Effect of the hyperparameters. (a)~Effect of varying weight $\beta_{cc}$. (b)~Influence of changing weight $\beta_{fa}$. (c)~Influence of varying weight $\beta_{ca}$. (d)~Effect of varying parameter $\varepsilon$.}
		\label{draw}
	\end{figure*}

	\subsection{Sensitivity Analysis}
	\label{section_parameter_analysis}
	There are mainly five hyperparameters in our model, including the weight of KD loss $\alpha$, the weight of category center contrast $\beta_{cc}$, the weight of feature alignment $\beta_{fa}$, the weight of category center alignment $\beta_{ca}$, and the growth rate $\varepsilon$. 
	$\alpha$ is set to $1$ for all experiments.
	For other three parameters, we conducted experiments on CIFAR-100 based on the architecture combinations of WRN-40-2 and WRN-16-2~\cite{zagoruyko2016wide} to analyze their sensitivity.
	The results are presented in Fig.~\ref{draw}. 
	
	\noindent \textbf{Effect of varying weight $\beta_{cc}$}. We fixed other weights as $\beta_{fa}=20$, $\beta_{ca}=1$, and made the magnitude of $\beta_{cc}$ vary in the range of $[0.01,0.1]$, increasing by $0.02$ each time. 
	According to Fig.~\ref{draw}~(a), we can see that our results are relatively stable. Even the minimum accuracy of $76.45\%$ is still much higher than other compared methods.
	
	\noindent \textbf{Influence of changing weight $\beta_{fa}$}. We fixed weights $\beta_{cc}=0.05$, $\beta_{ca}=1$ and let $\beta_{fa}$ range from $10$ to $50$ with a step size of $10$. 
	According to Fig.~\ref{draw}~(b), we observed that even for such a large range, the accuracy change is still acceptable, demonstrating that our method is insensitive to this parameter.
	
	\noindent \textbf{Influence of varying weight $\beta_{ca}$}. Similarly, we fixed weights $\beta_{cc}=0.05$, $\beta_{fa}=20$, and ranged $\beta_{ca}$ from $0.1$ to $10$. 
	According to Fig.~\ref{draw}~(c), we can see that even for such a large range, the difference between the best and the lowest results is less than $0.5\%$, indicating our method is also insensitive to this parameter. 
	
	\noindent \textbf{Effect of growth rate $\varepsilon$}. $\varepsilon$ controls the dynamic threshold of distinguishing difficult and easy samples. If this parameter is very large or small, it will not conducive to preview learning. So we set the range of $\varepsilon$ from $1.003$ to $1.05$, which ensures that the growth rate is within a reasonable range. 
	As shown in Fig.~\ref{draw}~(d), our model is also robust to this parameter.

    \begin{table*}[!t]
	    \renewcommand\arraystretch{1.05}
        
        \footnotesize
		\caption{Comparison between $\mathcal{L}_{CE}$ and preview strategy combined with $\mathcal{L}_{CE}$ on CIFAR-100.~({\color{red}{$\downarrow n$}}) represents a decrease of n. }
		\centering
		\begin{tabular}{c|ccccccc}
			\toprule
			Methods & WRN-16-2&  WRN-40-1  &ResNet32 & ResNet8x4  &  VGG8& ShuffleNetV1 & ShuffleNetV2   \\
			\midrule  
			$\mathcal{L}_{CE}$  & 73.71 & 71.38&  71.31 & 73.08 & 70.76  & 71.76 &72.93 \\
			$\mathcal{L}_{CE}+Preview$ 
            & 73.67({\color{red}{\scriptsize{$\downarrow$}\tiny{0.04}}}) 
            & 71.26({\color{red}{\scriptsize{$\downarrow$}\tiny{0.12}}}) 
            & 70.96({\color{red}{\scriptsize{$\downarrow$}\tiny{0.35}}}) 
            & 72.70({\color{red}{\scriptsize{$\downarrow$}\tiny{0.38}}}) 
            & 70.72({\color{red}{\scriptsize{$\downarrow$}\tiny{0.04}}}) 
            & 71.68({\color{red}{\scriptsize{$\downarrow$}\tiny{0.08}}}) 
            & 72.38({\color{red}{\scriptsize{$\downarrow$}\tiny{0.55}}})\\

			\bottomrule
		\end{tabular}
		\label{Preview_ce}
	\end{table*}

    \subsection{Analysis of Preview-based Learning Strategy} \label{ Preview-based}

     In this subsection, we first analyzed which loss should be combined with preview strategy, then we conducted a comparative evaluation between preview strategy and focal loss~\cite{DBLP:conf/iccv/LinGGHD17}, and finally, we compared preview strategy with classic curriculum learning strategy.

     \begin{figure}[!t]
		\centering
		\includegraphics[width=0.38 \textwidth]
        {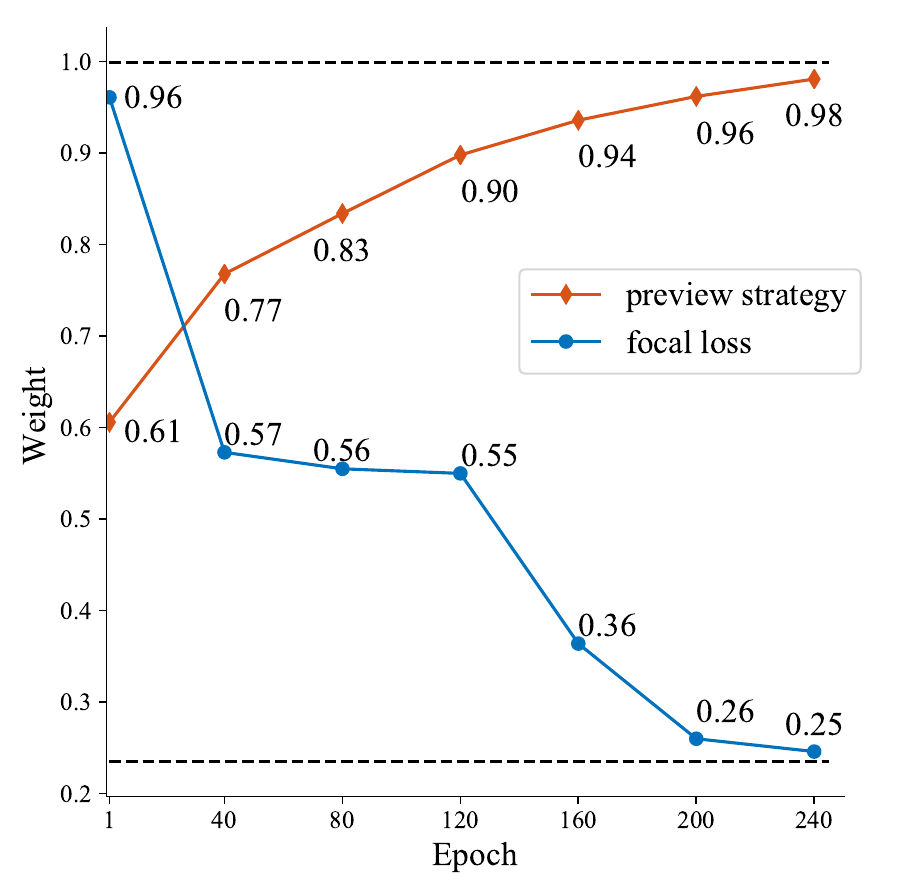}
		\caption{The average weights of all samples with preview strategy and focal loss as training progresses. }
		\label{focal_figure}
	\end{figure}

 %    \begin{table*}[htbp]
	%     \renewcommand\arraystretch{1.05}
	% 	\caption{Comparison between preview strategy and classic curriculum learning on CIFAR-100.}
	% 	\centering
	% 	\begin{tabular}{ccccccccc}
	% 		\toprule
	% 		Teacher & WRN-40-2& ResNet110& ResNet110& ResNet32x4 & VGG13 & ResNet50 & ResNet50 & ResNet32x4 \\
	% 		Student & WRN-16-2& ResNet20& ResNet32& ResNet8x4& MobileNetV2& MobileNetV2& VGG8 & ShuffleNetV1  \\
	% 		\midrule 
	% 		Curriculum-CKD& 76.41& 72.35 & 74.54&76.40 & 72.12 & 71.67 & 75.72 &78.47  \\
			
	% 		PCKD & \textbf{76.97}& \textbf{72.55} & \textbf{75.12}&\textbf{76.89} & \textbf{72.37}& \textbf{71.85} & \textbf{76.32} &\textbf{79.21} \\
	% 		\bottomrule
	% 	\end{tabular}
	% 	\label{t5}
	% \end{table*}
     \begin{table}[htbp] 
	    \renewcommand\arraystretch{1.05}
        \setlength{\tabcolsep}{4pt}
        \footnotesize 
		\caption{Comparison between preview strategy and focal loss on CIFAR-100. $*$ is focal loss and $\dag$ is preview strategy. }\centering 
		\begin{tabular}{lccc}
			\toprule
			\multirow{2}{*}{Methods}  & WRN-40-2 & ResNet110& WRN-40-2\\
			&   WRN-16-2 & ResNet20& ShuffleNetV1\\
			\midrule
			$\mathcal{L}_{KD}$ &74.84& 70.76& 75.37 \\

            $\mathcal{L}_{KD}*$  & 
            74.44 & 70.73 & 73.28 \\
			
			$\mathcal{L}_{KD}\dag$ & 75.08& 71.02& 75.48 \\
			
			$\mathcal{L}_{CKD}$ &  76.31& 72.04& 78.10 \\

            $\mathcal{L}_{CKD}*$ &  76.40&71.26 & 77.55 \\

            $\mathcal{L}_{CKD}\dag$ &  76.52& 72.05& 78.22 \\
        
            $\mathcal{L}_{KD}+\mathcal{L}_{CKD}$  & 76.42& 72.15& 78.15 \\
            $\mathcal{L}_{KD}*+\mathcal{L}_{CKD}*$
              & 76.83& 71.77& 77.41 \\
			
            $\mathcal{L}_{KD}\dag+\mathcal{L}_{CKD}\dag$  &\textbf{76.97}& \textbf{72.55}& \textbf{78.44} \\
			
			\bottomrule
		\end{tabular}
		\label{preview_focal}
    \end{table}

    \textbf{Which loss should be combined with preview strategy?} Knowledge distillation has multiple losses and we discussed below which loss should be combined with preview-based learning strategy. Cross-entropy loss $\mathcal{L}_{CE}$ learns whether the sample belongs to this category by judging whether the sample is the same as the sample of this category. If the weights are changed rashly, it may cause the classification to be biased towards the category with larger weights. 
    In order to prove our conjecture, we adopted preview strategy combined with cross-entropy loss to train various network structures on CIFAR-100. 
    As shown in Table~\ref{Preview_ce}, when cross-entropy loss is combined with preview strategy, the accuracy generally decreases, indicating that sample classification is biased towards certain categories, so we should not combine preview strategy with cross-entropy loss.
    
    Different from cross-entropy loss, $\mathcal{L}_{KD}$ and $\mathcal{L}_{CC}$ learn which category this sample belongs to. The model not only needs to determine whether the sample is in this category, but also should learn the probability value of the sample in other similar categories, i.e., \textit{dark knowledge}. Therefore, we proposed preview strategy to assign weights, which can help $\mathcal{L}_{KD}$ and $\mathcal{L}_{CC}$ to ignore the learning of hard samples at the beginning and focus on learning easy samples taught by the teacher. We demonstrated the effectiveness of this through ablation experiments in Table~\ref{tab:ablation}. $\mathcal{L}_{FA}$ and $\mathcal{L}_{CA}$ respectively align feature and category center to maintain architecture and parameter consistency between the teacher and student networks, without the need to allocate weights to focus on the difficulty of samples.

    \begin{figure*}[!t]
	\centering
	\includegraphics[width=1.02\textwidth]{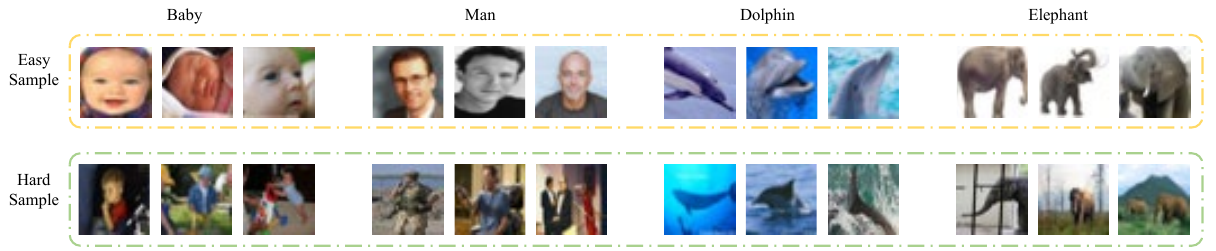}%
	\caption{Visualization of some easy and hard samples selected by our method based on results of the $50$th epoch under WRN-16-2 on CIFAR-100.}
	\label{hard_easy}
	\end{figure*}

	\begin{figure*}[!t]
		\centering
		\includegraphics[width=1.02\textwidth]{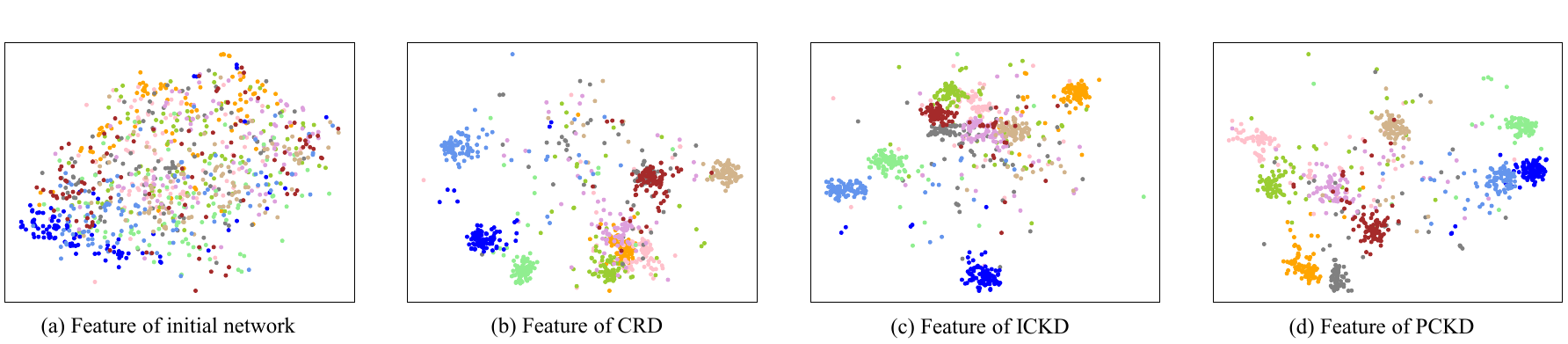}%
		\caption{We show the original feature from the initial network in~(a)~and compare the learned feature of t-SNE visualizations of CRD~\cite{tian2019contrastive} in~(b), ICKD~\cite{liu2021exploring} in~(c), and our proposed PCKD in~(d). Each color represents one category, a total of 10 categories.}
		\label{t-sne}
	\end{figure*}

    \textbf{Preview strategy \textit{vs} focal loss.} In order to prove the superiority of preview strategy, we compared it with focal loss~\cite{DBLP:conf/iccv/LinGGHD17}, which is also a dynamically weighting strategy. For better understanding of the distinction between preview strategy and focal loss, we used two strategies to train the WRN-16-2 network on CIFAR-100 dataset, and counted the weights of each training epochs. As shown in Fig.~\ref{focal_figure}, preview strategy gradually increases the weights of hard samples as training progresses. In contrast, focal loss exhibits an opposite trend, concentrating on hard samples to the extent of diminishing the weight assigned to easy ones. We chose the same network combination in subsection~\ref{Ablation Study} to conduct experiments on CIFAR-100 and compare the accuracy of two strategies.
    % The hyperparameters optimization of focal loss is shown in appendix~\ref{secA1}.
    As shown in Table~\ref{preview_focal}, we can see that adding focal loss is worse than preview strategy, proving that our preview strategy is better than focal loss in the field of image classification.  
    The reason is that the student model can better adapt to those hard samples in a progressive way instead of focusing exclusively on hard samples.

    \begin{table}[htbp]
	    \renewcommand\arraystretch{1.05}
		\caption{Comparison between preview strategy and classic curriculum learning on CIFAR-100.}
		\centering
		\begin{tabular}{cccc}
			\toprule
                Teacher & Student & Curriculum-CKD& PCKD \\
                \midrule 
			 WRN-40-2& WRN-16-2& 76.41 & \textbf{76.97} \\
              ResNet110& ResNet20& 72.35 & \textbf{72.55} \\
              ResNet110& ResNet32& 74.54& \textbf{75.12}\\
              ResNet32x4 & ResNet8x4& 76.40 & \textbf{76.89}\\
              VGG13 & MobileNetV2& 72.12 & \textbf{72.37}\\
              ResNet50 & MobileNetV2& 71.67 & \textbf{71.85}\\
              ResNet50 & VGG8 & 75.72 & \textbf{76.32} \\
              ResNet32x4 & ShuffleNetV1 & 78.47 & \textbf{79.21}\\
			    
			\bottomrule
		\end{tabular}
		\label{t5}
	\end{table}
    
    \textbf{Preview strategy \textit{vs} curriculum learning strategy.} To further validate the advantage of our preview strategy over the classic curriculum learning strategy, we conducted additional experiments on CIFAR-100 under eight architecture combinations, and the results are shown in Table~\ref{t5}. Curriculum-CKD denotes the method that replaces preview strategy with curriculum learning, where those hard samples are directly filtered out in each mini-batch, so we set weights of hard samples to $0$ in Eq.~(\ref{e6}). We can also observe that PCKD achieves much better results than Curriculum-CKD, showing our superiority. 
	This is attributed to the student model's enhanced ability to adapt progressively to challenging samples through preview strategy.

    \subsection{Visualization}
    We first showed some easy and hard examples of several categories distinguished by our preview strategy in Fig.~\ref{hard_easy}. It is obvious that the difference between easy and hard samples is very apparent, and the hard samples are really complicated to recognize at first glance. In this case, it is necessary to adopt our preview strategy.
	
	To demonstrate the superiority of our category contrastive learning strategy in discriminative feature learning, we used t-SNE~\cite{maaten2008visualizing} to visualize the feature representation of the student network on CIFAR-100. VGG13 and VGG8 are selected as the teacher and the student networks, respectively. We chose the feature at the penultimate layer and randomly selected $10$ categories from $100$ categories for visualization. The results are shown in Fig.~\ref{t-sne}. We can observe that CRD~\cite{tian2019contrastive} and ICKD~\cite{liu2021exploring} are indistinguishable in some categories. Specifically, features of two categories in CRD are completely overlapped, and 
	features of three categories in ICKD are very close. 
	In contrast, features of different categories in PCKD are very distinct, which demonstrates that our method can learn more discriminative feature representation, leading to better classification results.

\section{Conclusion}\label{d}
	In this paper, we propose a novel preview-based category contrastive learning method for knowledge distillation. On the one hand, PCKD transfers the structural knowledge of the feature representation, the category representation, and the relation between them in a contrastive learning framework, which can learn more discriminative features and category centers, leading to better results. 
	On the other hand, our method can train the student network according to the difficulty of samples by adopting a preview strategy based on curriculum learning. 
	Sample-specific weights are dynamically learned to guide the network training in a progressive way. We also conduct extensive experiments on several challenging datasets under various combination of teacher and student networks, and our method achieves superior results consistently.
	For future work, we would like to extend it to the self-supervised KD and  multi-teachers guided KD, which can further improve the performance of the student network.

\bibliographystyle{IEEEtran}
	%\bibliography{reference_list}
    \bibliography{PCKD}

    \begin{IEEEbiography}[{\includegraphics[width=1in,height=1.25in,clip,keepaspectratio]{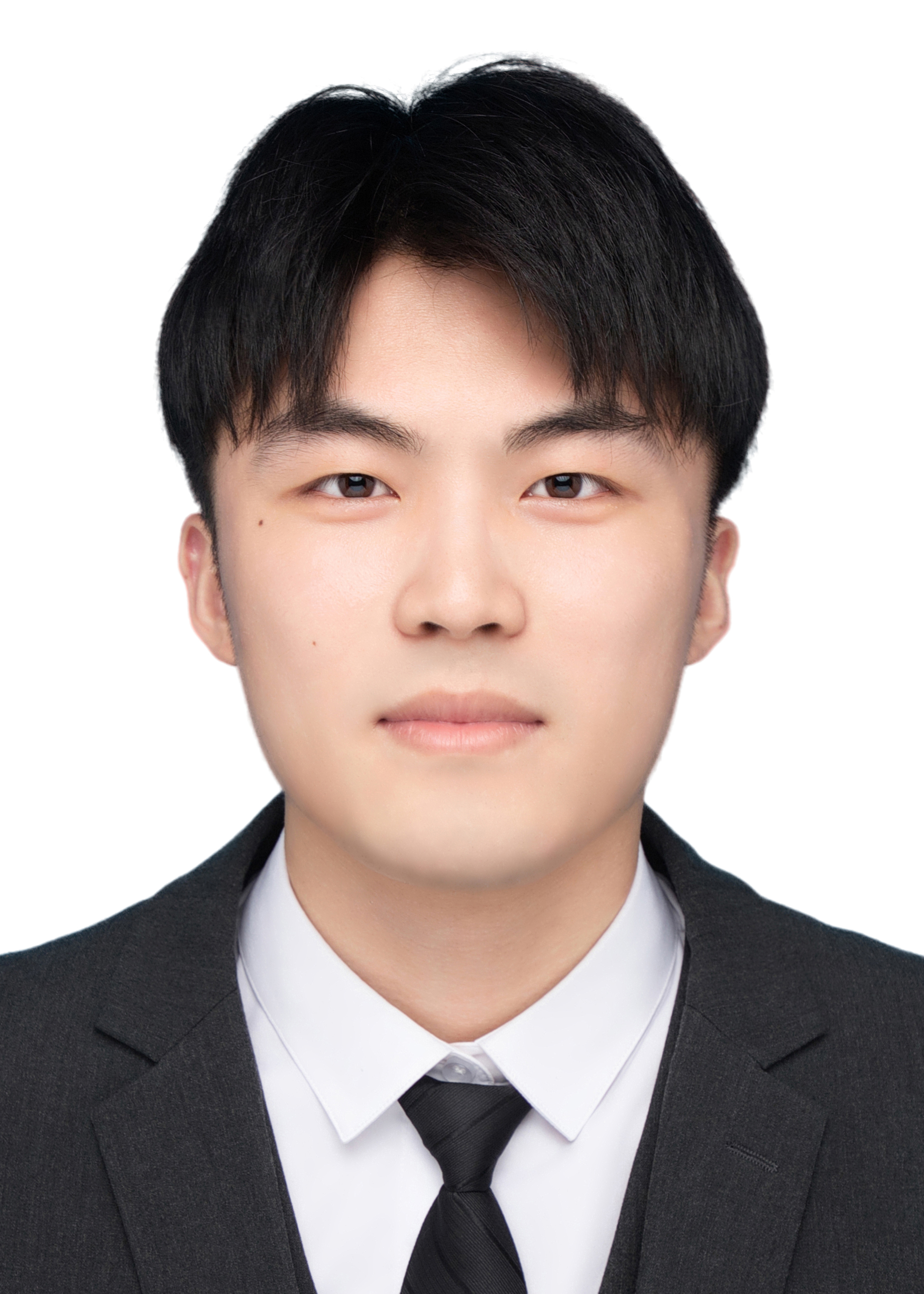}}]{Muhe Ding} 
		received the bachelor's degree in computer science from Shandong University in 2021. He is currently a master student in the School of Computer Science and Technology, Shandong University. His research interests contain computer vision and multimodal learning.
	\end{IEEEbiography}
	
	\begin{IEEEbiography}[{\includegraphics[width=1in,height=1.25in,clip,keepaspectratio]{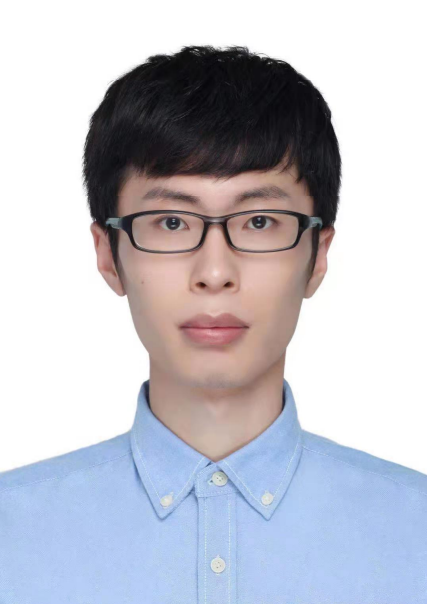}}]{Jianlong Wu}(Member, IEEE) 
        received his B.Eng. and Ph.D. degree from Huazhong University of Science and Technology in 2014 and Peking University in 2019, respectively. He is currently an associate professor with Harbin Institute of Technology (Shenzhen). His research interests lie primarily in computer vision and machine learning. He has published more than 30 papers in top journals and conferences, such as TIP, ICML, NeurIPS, and ICCV. He serves as a Senior Program Committee Member of IJCAI 2021, an area chair of ICPR 2022/2020, and a reviewer for many top journals and conferences, including TPAMI, IJCV, ICML, and CVPR. He received many awards, such as the outstanding reviewer of ICML 2020, and the Best Student Paper of SIGIR 2021.
	\end{IEEEbiography}

	\begin{IEEEbiography}[{\includegraphics[width=1in,height=1.25in,clip,keepaspectratio]{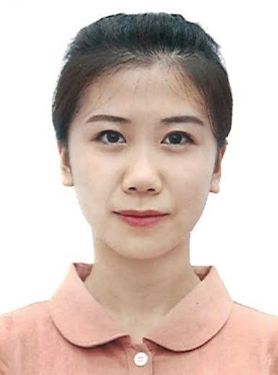}}]{Xue Dong} received the B.E. degree from School of Mathematical Sciences in 2014, University of Jinan, China. She is currently a Ph.D. candidate of School of Software, Shandong University, China. She has published several papers in the top venues.
		Her research interests contain multimedia computing, retrieval and recommendation.
	\end{IEEEbiography}
	
	\begin{IEEEbiography}[{\includegraphics[width=1in,height=1.25in,clip,keepaspectratio]{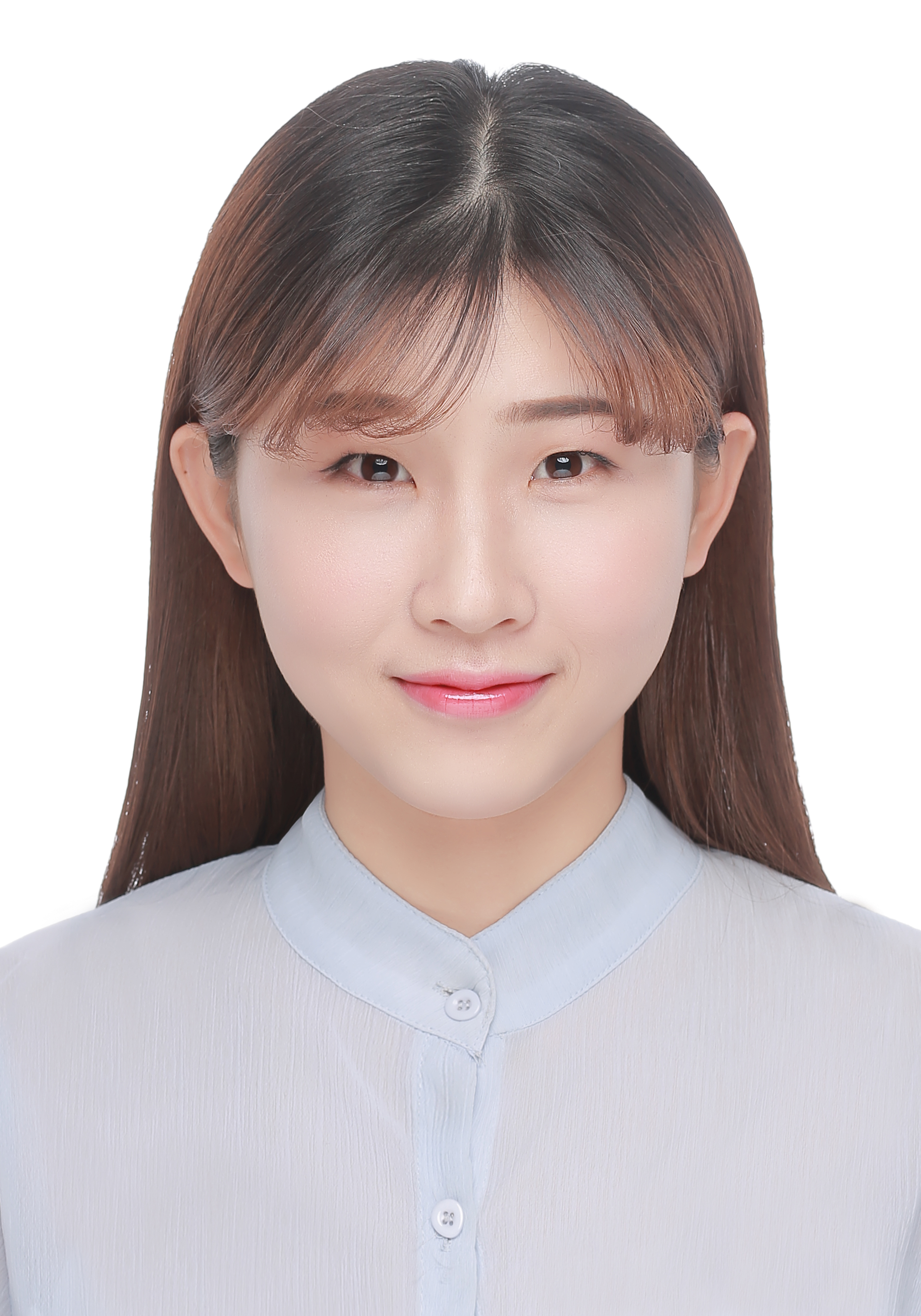}}]{Xiaojie Li} 
		is a PhD candidate in Computer Science and Technology at Harbin Institute of Technology (Shenzhen), earned her Bachelor's and Master's degrees from Beihang University's School of Instrumentation and Optoelectronic Engineering in 2016 and 2019, respectively. Her research focuses on computer vision, self-supervised learning, knowledge distillation, and multimodal learning.
	\end{IEEEbiography}
	
	\begin{IEEEbiography}[{\includegraphics[width=1in,height=1.25in,clip,keepaspectratio]{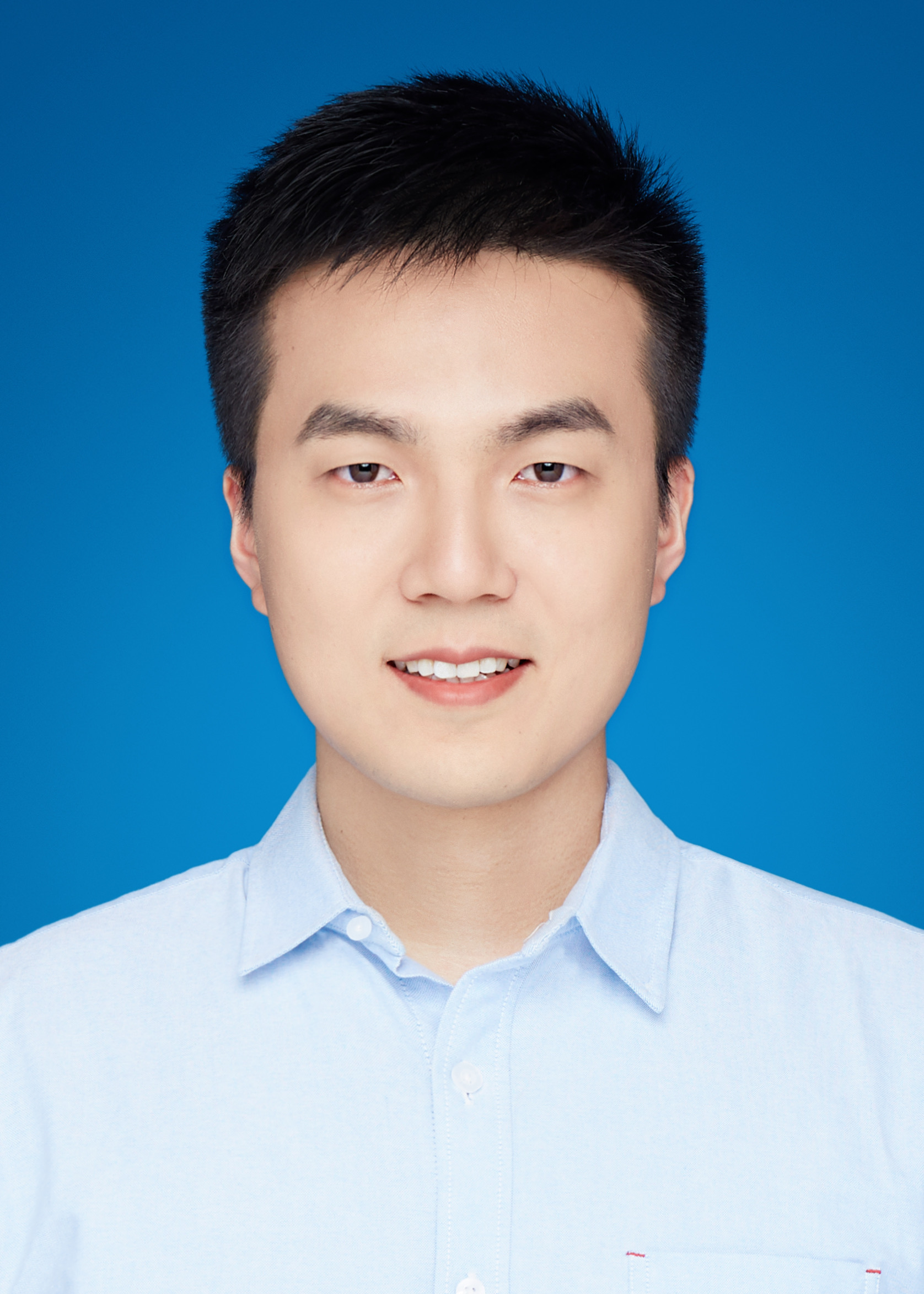}}]{Pengda Qin} received the Ph.D. degree from Beijing University of Posts and Telecommunications in 2020. He is currently a senior engineer at Alibaba Group. His research interests contain Natural Language Processing, Self-Supervised Learning, Large Multimodal Model.
	\end{IEEEbiography}
	
	\begin{IEEEbiography}[{\includegraphics[width=1in,height=1.25in,clip,keepaspectratio]{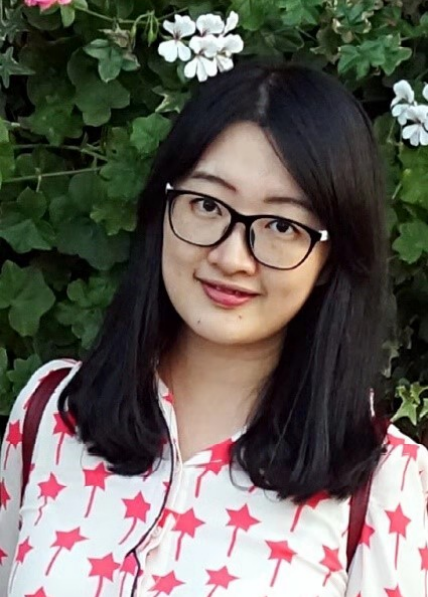}}]{Tian Gan} 
		is currently an Associate Professor with the School of Computer Science and Technology, Shandong University. She received her B.S. from East China Normal University in 2010, and the Ph.D. degree from National University of Singapore, Singapore, in 2015. She was a Research Scientist in Institute for Infocomm Research (I2R), Agency for Science, Technology and Research (A*STAR). Her research interests include social media marketing, video understanding, and multimedia computing.
	\end{IEEEbiography}
	
	\begin{IEEEbiography}[{\includegraphics[width=1in,height=1.25in,clip,keepaspectratio]{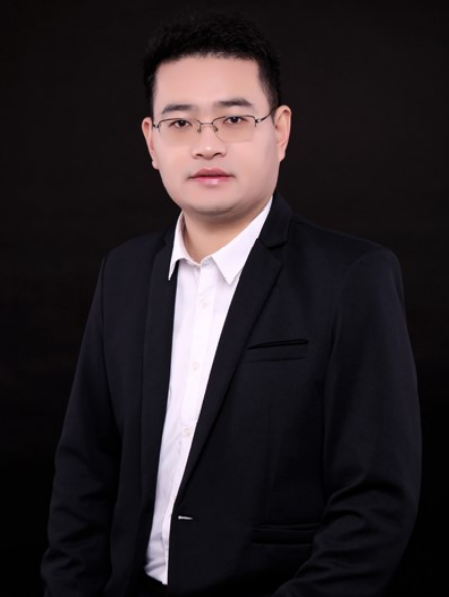}}]{Liqiang Nie} (Senior Member, IEEE) 
		 is currently the dean with the Department of Computer Science and Technology, Harbin Institute of Technology (Shenzhen). He received his B.Eng. and Ph.D. degree from Xi’an Jiaotong University and National University of Singapore (NUS), respectively. His research interests lie primarily in multimedia computing and information retrieval. Dr. Nie has co-/authored more than 100 papers and 4 books, received more than 15,000 Google Scholar citations. He is an AE of IEEE TKDE, IEEE TMM, IEEE TCSVT, ACM ToMM, and Information Science. Meanwhile, he is the regular area chair of ACM MM, NeurIPS, IJCAI and AAAI. He is a member of ICME steering committee. He has received many awards, like ACM MM and SIGIR best paper honorable mention in 2019, SIGMM rising star in 2020, TR35 China 2020, DAMO Academy Young Fellow in 2020, SIGIR best student paper in 2021, ACM MM best paper in 2022.
	\end{IEEEbiography}

\end{document}